\documentclass[twoside]{article}

\usepackage[accepted]{aistats2023}
\usepackage{subfiles}

\usepackage[utf8]{inputenc} 
\usepackage[T1]{fontenc}    
\usepackage{graphicx}
\usepackage{hyperref}       
\usepackage{url}            
\usepackage{booktabs}       
\usepackage{natbib}
\usepackage{amsthm}
\usepackage{enumitem}

\usepackage{amsfonts}       
\usepackage{bbm}
\usepackage{nicefrac}       

\usepackage{xcolor}         
\usepackage[mathcal]{eucal}
\newtheorem{theorem}{Theorem}

\theoremstyle{remark}

\DeclareMathAlphabet\mathbfcal{OMS}{cmsy}{b}{n}
\newcommand{\BEAS}{\begin{eqnarray*}}
\newcommand{\EEAS}{\end{eqnarray*}}
\newcommand{\BEA}{\begin{eqnarray}}
\newcommand{\EEA}{\end{eqnarray}}
\newcommand{\BEQ}{\begin{equation}}
\newcommand{\EEQ}{\end{equation}}
\newcommand{\BIT}{\begin{itemize}}
\newcommand{\EIT}{\end{itemize}}
\newcommand{\BNUM}{\begin{enumerate}}
\newcommand{\ENUM}{\end{enumerate}}
\newcommand{\BA}{\begin{array}}
\newcommand{\EA}{\end{array}}
\newcommand{\diag}{\mathop{\rm diag}}

\newcommand{\tr}{\mathop{ \rm tr}}

\newcommand{\idm}{I}
\newcommand{\rb}{\mathbb{{R}}}

\newcommand{\ds}{\displaystyle }

 \def \ds { \displaystyle}

\def \E{{\mathbb E}}

\newcommand{\punt}[1]{}

\theoremstyle{plain}

\newcommand{\reals}{\mathbb{R}}

\newcommand{\bq}{\begin{equation}}
\newcommand{\ba}{\begin{eqnarray}}
\newcommand{\ea}{\end{eqnarray}}

\def\R{{\reals}}

\newcommand{\remove}[1]{}

\begin{document}

%
\runningtitle{Explicit Regularization in Overparametrized Models via Noise Injection}

%
\runningauthor{Antonio Orvieto$^*$, Anant Raj$^*$, Hans Kersting$^*$ and Francis Bach}

\twocolumn[

\aistatstitle{Explicit Regularization in Overparametrized Models via Noise Injection}

\aistatsauthor{%
  Antonio Orvieto$^*$
  \And 
  Anant Raj$^*$
}


\aistatsaddress{   Department of Computer Science\\
  ETH Zürich, Zürich, Switzerland\\
  \And  
   Coordinated Science Laboraotry \\
  University of Illinois Urbana-Champaign \\
  Inria, Ecole Normale Sup\'erieure \\
  PSL Research University, Paris, France}

\aistatsauthor{  Hans Kersting$^*$
  \And 
  Francis Bach }

\aistatsaddress{Inria, Ecole Normale Sup\'erieure \\
  PSL Research University, Paris, France\\
  \And  Inria, Ecole Normale Sup\'erieure \\
  PSL Research University, Paris, France }]

\begin{abstract}
\vspace{-2mm}
Injecting noise within gradient descent has several desirable features, such as smoothing and regularizing properties. In this paper, we investigate the effects of injecting noise before computing a gradient step. We demonstrate that small perturbations can induce explicit regularization for simple models based on the $\ell_1$-norm, group $\ell_1$-norms, or nuclear norms. However, when applied to overparametrized neural networks with large widths, we show that the same perturbations can cause variance explosion. To overcome this, we propose using independent layer-wise perturbations, which provably allow for explicit regularization without variance explosion. Our empirical results show that these small perturbations lead to improved generalization performance compared to vanilla gradient descent.
\end{abstract}
\vspace{-2mm}

\section{Introduction}
Injecting noise  in gradient descent has several desirable properties. In previous works \citep{jin2017escape,reddi2018generic,staib2019escaping},  noise injection \emph{after the gradient step} has been used to escape saddle points efficiently in non-convex optimization. In several recent works \citep{zhu2018anisotropic,nguyen2019first}, it was also shown that noise injection in  gradient descent helps to escape local minima while optimizing non-convex functions.  Adding noise \emph{within the gradient}, either on labels or on the iterates before computing the (stochastic) gradient, has  been classically used to smooth the objective function (often in the convex case) \citep{nesterov2017random,polyak1990optimal,flaxman2004online,duchi2012randomized,d2014stochastic,nesterov2005smooth}. However, the effect of noise injection on generalization is not as well understood. 

Recently, there has been some interests in understanding the geometry of the loss landscape enforced by the induced noise and its connection with generalization.  A recent result \citep{liu2021noisy} shows that perturbed gradient descent with sufficiently small noise converges to flat minima for non-convex matrix factorization tasks. \citet{orvieto2022anticorrelated} also provide some explanation and empirical evidence for better performance of noise induced gradient descent as the noise injection favors flat minima.  

In this work, we explore several avenues along this  line of research. Our aim is to tackle the following questions:
\begin{enumerate}[leftmargin=0.42cm,topsep=1pt,itemsep=2pt,partopsep=0.5pt, parsep=0pt]
\item Do small perturbations in the model while training induce an explicit form of regularization for simple   models common in machine learning? 

\item 	Do the same  perturbations also induce an explicit regularization effect  while training overparametrized models such as homogeneous neural networks with large (infinite) widths? 

\item Is the induced regularization useful in practice, i.e., does it lead to better generalization performance?
\end{enumerate}

In response to the questions above, we make the following contributions: 
\begin{enumerate}[leftmargin=0.48cm,topsep=1pt,itemsep=2pt,partopsep=0.5pt, parsep=0pt]
\item We show in Section~\ref{sec:finitedim} that small perturbations induce explicit regularization for simple  models, and that this induced regularization is based on the $\ell_1$-norm, group $\ell_1$-norm, or nuclear norms. Such explicit formulations were available before for only special cases ($\ell_1$-norm), or for special loss functions (logistic regression in the separable case).

\item We show in Section~\ref{sec:neural-net} that the same perturbations do not work for overparametrized neural networks with large widths due to variance explosion resulting from the number of perturbed neurons tending to infinity. However, we also show that \emph{independent layer-wise perturbations} (where only one layer is perturbed at each gradient step)  allow to avoid the exploding variance term, and explicit regularizers can be obtained, for a fully connected network with a single hidden layer. This is extended in Section~\ref{sec:neural-net-deep} to deep ReLU networks.

\item We empirically show in Section~\ref{sec:exp} that small layer-wise perturbations lead to better generalization performance in practice over vanilla (stochastic) gradient descent training, for a variety of shallow and deep overparametrized models, fully connected or convolutional, with minor adjustments to the training procedure.
\end{enumerate}

\subsection{Related work} 
\label{sec:related_work}
Noise injection has been frequently used for various tasks in statistics, machine learning and  signal processing. In this section, we discuss three  major directions of research and existing literature associated with them which are closely connected to our work. 

\vspace{-3mm}
\paragraph{Perturbed gradient descent (PGD).}
PGD methods are versions of (stochastic) gradient descent (GD) where noise is injected in one of the following two ways: In the first option, perturbations are added \emph{after} each GD step.
Such versions of PGD have been shown to speed up the escape from spurious local minima and saddle points; see \citet{zhou2019pgd} and \citet{Jin2021pgd}, respectively.
These methods correspond to discretizations of the continuous-time Langevin dynamics \citep{Li2017SMEs} where the perturbations add to the diffusion coefficient. In the second option, perturbations are added to the iterate \emph{before} the gradient is evaluated. 
The so-perturbed gradient is then used to update the unperturbed iterate. In the convex case, this is typically used to smooth a non-smooth objective function~\citep{nesterov2017random,polyak1990optimal,flaxman2004online,duchi2012randomized,d2014stochastic,nesterov2005smooth}. In the non-convex case, such versions of PGD have been shown to exhibit an implicit bias toward flat minima (that generalize better) in the case of nonconvex matrix factorization \citep{liu2021noisy} and more general models \citep{orvieto2022anticorrelated}. Flat minima are widely believed to have a better generalization property \citep{keskar2016large,chaudhari2019entropy,jiang2019fantastic}, and we follow this line of work, with several generalization guarantees that are based on PAC-Bayesian analysis and also apply to our framework~\citep{tsuzuku2020normalized,neyshabur2017pac}.

The second option of PGD can be interpreted as an instance of the first option where the perturbations are instead anticorrelated.
Such methods are therefore also referred to as \emph{Anti-PGD}; see Section 2 from \citet{orvieto2022anticorrelated}.
In our current paper, we determine the \emph{effective loss} that is (on average) used in the second option. 

\vspace{-3mm}

\paragraph{Explicit and implicit regularization in learning.} It is   well understood   that regularization plays a crucial role in the generalization performance of a learning model. The most common form of regularization is Tikhonov regularization \citep{weese1993regularization,golub1999tikhonov}, which is imposed explicitly in the optimization objective for linear and non-linear models.  There are other ways to induce regularization explicitly while training machine learning models \citep{hanson1988comparing,srivastava2014dropout,raj2021explicit}. Apart from explicit regularization methods, optimization algorithms can also induce implicit bias on the optimal solution, which was observed by \citet{soudry2018implicit,gunasekar2018characterizing,kubo2019implicit}. In a closely related line of work, \citet{camuto2020explicit}  study the regularization induced in neural networks by Gaussian noise injections to all 
network activations, and studies its regularization effect in the Fourier domain. However, while they obtain explicit regularizers similar to ours, we study the effect of noise injection to the model parameter instead of network activations, and hence our analysis is applicable to general machine learning models. Moreover, we deal with infinite widths through layer-wise perturbations.

\vspace{-3mm}

\paragraph{Robustness.} Apart from inducing regularization and finding flatter minima, noise injection is also helpful in getting robust models. Recent works \citep{salman2019provably,cohen2019certified,lecuyer2019certified,li2018second} have shown that randomized smoothing is a scalable way of building provably robust neural network based classifiers. However, these approaches rely on  input perturbation instead of model perturbation.  \citet{wu2020adversarial} showed  that adversarial weight perturbation helps in flattening the weight loss landscape, which improves the  robust generalization gap. \citet{zheng2021regularizing} also employ adversarial model perturbation to  to favor flat local minima of the empirical risk. However, a systematic study of the influence of noise injection in the model is lacking.


\vspace{-2mm}

\section{Finite-dimensional models}
\label{sec:finitedim}
\vspace{-2mm}
In this section, we consider a loss function $L: \rb^n \to \rb$, as well as a predictor $\Phi: \rb^m \to \rb^n$, and we aim to minimize
\vspace{-2mm}
\begin{equation}
    R(w) = L(\Phi(w))
\end{equation}
with respect to $w \in \rb^m$. In machine learning contexts, $\Phi(w)$ represents the $n$ real-valued predictions over $n$ inputs, and $L$ is the associated loss function, typically, $L(\varphi) = \frac{1}{2n} \sum_{i=1}^n ( y_i - \varphi_i)^2$ for least-squares regression, or $L(\varphi) = \frac{1}{n} \sum_{i=1}^n \log (1 + \exp(- y_i \varphi_i))$ for logistic regression, where $y \in \rb^n$ is the vector of output labels. The framework applies equally well to multi-dimensional outputs (such as multi-category classification and multinomial loss), where $\rb^n$ is replaced by $\rb^{n \times k}$ (this is in fact considered in Sections~\ref{sec:linear_networks},~\ref{sec:neural-net} and~\ref{sec:neural-net-deep} for neural networks).

In this section, we only consider the asymptotics with respect to $\sigma$, assuming that the number of parameters $m$ remains fixed (hence the denomination ``finite-dimensional''). We analyze the effect of strong overparametrization with the number of parameters tending to infinity in Sec.~\ref{sec:neural-net} and~\ref{sec:neural-net-deep}.

\subsection{Gaussian perturbations and Taylor expansions}
Following previous work on PGD (see Section~\ref{sec:related_work}), we consider perturbing the iterate by a standard Gaussian vector $\varepsilon$ (with zero mean and identity covariance matrix), and consider the function
\begin{equation} \label{eq:def_Rsigma}
    R_\sigma(w) = \E \big[
    L(\Phi(w+ \sigma \varepsilon))
    \big].
\end{equation}
We denote by ${\rm D}\Phi(w) \in \rb^{n \times m}$ the Jacobian (matrix of first-order derivatives) of $\Phi$, and ${\rm D}^2\Phi(w) \in \rb^{ n \times m \times m}$ the tensor of second-order derivatives.
\vspace{-2mm}
\paragraph{Taylor expansions.} Throughout this section, we assume for simplicity of exposition that $L$ and~$\Phi$ are three times differentiable functions with bounded third derivatives. In Appendix~\ref{ap:proof_theorems}, we discuss the case of functions where this is satisfied only piece-wise (to cover ReLU activations).

The Taylor expansion of $\Phi$ at $w$ can be written:
\begin{align}
&\textstyle \Phi(w + \sigma \varepsilon)\nonumber = \Phi(w) + \sigma {\rm D}\Phi(w) \varepsilon\\ & \qquad \qquad \qquad  + \frac{\sigma^2}{2} {\rm D}^2 \Phi(w) [ \varepsilon \varepsilon^\top] + O(\sigma^3 \| \varepsilon\|^3),\label{eq:taylor_1}
\end{align}
where  $\ds  {\rm D}^2 \Phi(w) [M ] \in \rb^n$ is defined as $  {\rm D}^2 \Phi(w) [M ]_j = \sum_{a,b =1}^m  {\rm D}^2 \Phi(w)_{jab} M_{ab}$. 

We also need a Taylor expansion of $L$ around $\Phi(w)$, which can be written as
\begin{align}
\textstyle
L(\Phi(w) + \Delta)
= L(\Phi(w)) + {\rm D} L(\Phi(w)) \Delta  \notag \\
+  \frac{1}{2}{\rm D^2} L(\Phi(w)) [ \Delta \Delta^\top ]
 + O ( \| \Delta \|^3), \label{eq:loss-exp} 
\end{align}
where, for $\varphi \in \rb^n$, ${\rm D} L(\varphi) \in \rb^{1 \times n}$ is the row-vector of first-order partial derivatives, and ${\rm D^2} L(\varphi) \in \rb^{n \times n}$ the matrix of second-order partial derivatives, with the notation ${\rm D^2} L(\varphi) [ M ] = \sum_{a,b=1}^n {\rm D^2} L(\varphi)_{ab} M_{ab} \in \rb$. 

To compute an expansion of $R_\sigma$, as defined in Eq.~\eqref{eq:def_Rsigma}, we compose the two expansions and get, with $\Delta = \sigma {\rm D}\Phi(w) \varepsilon + \frac{\sigma^2}{2} {\rm D}^2 \Phi(w) [ \varepsilon \varepsilon^\top] + O(\sigma^3 \| \varepsilon\|^3)$ the expansion
\small
\begin{align*}
 &L(\Phi(w+\sigma \varepsilon)) \\
 &=  \textstyle L \big(
 \Phi(w) + \sigma {\rm D}\Phi(w) \varepsilon + \frac{\sigma^2}{2} {\rm D}^2 \Phi(w) [ \varepsilon \varepsilon^\top] + O(\sigma^3 \| \varepsilon\|^3)\big) \\
 &= \textstyle L (
 \Phi(w))  + DL(\Phi(w)) \big( \sigma {\rm D}\Phi(w) \varepsilon + \frac{\sigma^2}{2} {\rm D}^2 \Phi(w) [ \varepsilon \varepsilon^\top]  \\
 & \qquad \textstyle + O(\sigma^3 \| \varepsilon\|^3)\big)  + 
 \frac{1}{2}{\rm D^2} L(\Phi(w)) \big[
 \sigma {\rm D}\Phi(w) \varepsilon (\sigma {\rm D}\Phi(w) \varepsilon)^\top \\
 & \hspace*{4cm}+ O(\sigma^3 \| \varepsilon\|^3) \big]
  + O(\sigma^3 \| \varepsilon\|^3) .
\end{align*}
\normalsize
Taking expectations, using $\E [ \varepsilon ] = 0$ and $\E [ \varepsilon \varepsilon^\top] = I$:
\begin{align}
  \textstyle  R_\sigma(w) 
   =  L (
 \Phi(w)) +   \frac{\sigma^2}{2} {\rm D}  L(\Phi(w)) {\rm D}^2 \Phi(w) [  \idm ] \notag \\
 + \frac{\sigma^2}{2} {\rm D^2} L(\Phi(w)) \big[ {\rm D}\Phi(w) {\rm D}\Phi(w)^\top \big] + O(\sigma^3). \label{eq:noise_taylor}
\end{align}
(The careful reader will have noticed that we assumed that $\Phi$ is in $C^3$; in Appendix~\ref{ap:proof_theorems} we present an extension to the case where $\Phi$ is only piecewise in $C^3$ --- e.g. ReLU.)
Note that the added terms on top of $R(w) = L(\Phi(w))$ can be derived as well from the trace of the Hessian of $R$, as obtained by~\citet{orvieto2022anticorrelated}.

\vspace{-1mm}
\textbf{Minimizing $R_\sigma$ with stochastic gradient descent.}  
As discussed by \citet{orvieto2022anticorrelated}, in order to minimize $R_\sigma$ defined as an expectation in Eq.~\eqref{eq:def_Rsigma}, we do not need to compute the expectation: we simply use a single perturbation (i.e., a single Gaussian vector $\varepsilon$), and compute the gradient with respect to parameters,   which leads to an unbiased estimate that can be used within a stochastic gradient descent algorithm. Since the loss function $L$ has a finite sum structure, we can also use stochastic gradient for over observations, as traditionally done in machine learning.
\vspace{-2mm}

\subsection{Asymptotically equivalent objective functions}

In several situations, we will show that we obtain the following ``asymptotically equivalent'' cost function $ R_\sigma^{({\rm eff})} (w)$, defined as
\begin{equation}
 \textstyle   R_\sigma^{({\rm eff})} (w) =  R(w) +  \frac{\sigma^2}{2} {\rm D^2} L(\Phi(w)) \big[ 
  {\rm D}\Phi(w) {\rm D}\Phi(w)^\top
 \big], \label{eq:second_order}
\end{equation}
in the sense that finding a global minimizer of $R_\sigma(w) = \E \big[
    L(\Phi(w+ \sigma \varepsilon))
    \big] $ is essentially equivalent to finding a global minimizer of $ R_\sigma^{({\rm eff})} (w)$ as $\sigma \to 0$.
    In other words, the term $ \frac{\sigma^2}{2} {\rm D}  L(\Phi(w)) {\rm D}^2 \Phi(w) [  \idm ]  $ in Eq.~(\ref{eq:noise_taylor}) has no impact. This will be done by showing that the order of approximation between $R$ and $R_\sigma$ is of order $\sigma^2$, while the one between $R_\sigma$ and $ R_\sigma^{({\rm eff})} $ is of order~$\sigma^3$, in two different cases.

\vspace{-2mm}

\paragraph{Hessians with no cross-products.}
One simple sufficient condition for 
the term $ \frac{\sigma^2}{2} {\rm D}  L(\Phi(w)) {\rm D}^2 \Phi(w) [  \idm ]  $ to have no impact 
is that ${\rm D}^2 \Phi(w) [  \idm ]=0$ for all $w \in \rb^m$. This is satisfied for certain models such as neural networks, where the second-order derivatives have no cross-product terms, or equivalently when $\tr(\Phi_j(w)) = 0$ for all $j=1,\dots,m$; this is e.g.~the case when the ReLU activation is used, see Sections~\ref{sec:linear_networks},~\ref{sec:neural-net}, and~\ref{sec:neural-net-deep} for more such examples. The following theorem makes it precise for sufficiently regular functions. See proof and more general results in App.~\ref{ap:proof_theorems} (already sketched above).

\begin{theorem} 
\label{theo:cross}
    Assume that (a) $\Phi$ and $L$ are three-times continuously differentiable with uniformly bounded third derivatives, (b) ${\rm D}^2 \Phi(w) [  \idm ]=0$ for all $w \in \rb^m$. Then, there exist constants $C$ and $C'$ (independent of $w$)
    such that   $\forall w \in \rb^m$,  $|R_\sigma(w) -  R(w)| \leq C(1 + \| w \|^2)\sigma^2$  while
    $|R_\sigma(w) -  R_\sigma^{({\rm eff})}(w)| \leq C'\sigma^3$.
 \end{theorem}

\vspace{-3mm}

\paragraph{Overparametrized models.}
A more general condition is related to overparametrization (when the model is sufficiently rich to lead to the minimizer of $L$), and we make a formal statement below with simplified assumptions. See Appendix~\ref{ap:proof_theorems} for the proof and more refined conditions.

\begin{theorem}
\label{theo:over}
Assume that (a) $\Phi$ is three-times continuously differentiable with uniformly bounded second and third derivatives,  (b) $L$ is strongly convex with uniformly bounded second and third derivatives, with unique global minimizer $\varphi^\ast$, (c) there exists a (non-unique) $w_\ast \in \rb^m$ such that $\Phi(w_\ast) = \varphi_\ast$,
and that (d) there exist minimizers
$w^{\sigma}_\ast$ and $w^{\sigma, (\rm eff)}_\ast$ of $R_\sigma$ and $R_\sigma^{({\rm eff})}$ that lie in a compact set $\Omega \subset \mathbb{R}^m$. Then, if 
$w^{\sigma}_\ast$ is a minimizer of $R_\sigma$,  and $w^{\sigma, (\rm eff)}_\ast$ is a minimizer of $R_\sigma^{({\rm eff})}$, we have $\|
\Phi(w^{\sigma}_\ast) -  \varphi_\ast \|_2^2 = O(\sigma^2)$
while 
$\|
\Phi(w^{\sigma, (\rm eff)}_\ast) -  \Phi(w^{\sigma}_\ast) \|_2^2 = O(\sigma^{3})$.
\end{theorem}
\vspace{-2mm}
Note that (a) we characterize the prediction function $\Phi$ taken at the various minimizers to characterize asymptotic equivalence (but it is not possible to characterize a distance between parameters because in overparametrized models, $\Phi$ cannot be injective), and (b) when $\sigma$ tends to zero the minimizer should converge to the interpolator $\Phi(w) = \varphi_\ast$ with minimal
${\rm D^2} L(\Phi(w)) \big[ 
  {\rm D}\Phi(w) {\rm D}\Phi(w)^\top
 \big]$.

Below, we consider specific cases for the least square loss and the logistic  loss in the separable case, before considering specific simple models for $\Phi$.

\textbf{Least-squares regression.}
We consider $L(\varphi) = \frac{1}{2n} \| y - \varphi\|_2^2$, which is  a quadratic function with constant  Hessian $\frac{1}{n} I$. We therefore get an extra regularizer on top of $R(w)$ equal to $\frac{\sigma^2}{n} \tr \big[  {\rm D}  L(\Phi(w)) {\rm D}  L(\Phi(w))^\top \big] $. \\
Note that our regularizer is then related to the neural tangent kernel~\citep{jacot2018neural}, i.e., this is the trace of the corresponding kernel matrix. While this could lead to interesting further connections, we note that our work applies beyond the lazy training regime (that is, our parameters $w$ can vary significantly)~\citep{chizat2019lazy}.

\textbf{Separable logistic regression.}
If we consider the logistic loss with an overparametrized model, then the infimum of $L$ is attained at infinity, and we should expect $\Phi(w)$ to diverge along a specific direction, like in other similar models~\citep{soudry2018implicit,chizat2020implicit}.\\
For homogeneous models like considered by~\citet{lyu2019gradient}, we can expect $w$ to diverge in some direction, that is, $w = \lambda \Delta$ for $\lambda \to +\infty$. The term ${\rm D}  \Phi(w) {\rm D}(\Phi(w))^\top$ grows  in $\lambda$, while ${\rm D}^2 L(\Phi(w))$ converges to zero exponentially fast. Overall, we  conjecture (and empirically observe) that we get divergence, and that the direction $\Delta$ ends up being proportional to the minimizer of
$\| {\rm D}  \Phi(\Delta) \|_F^2$ such that $ y \circ \Phi(\Delta) \geq  1$, where $y \in \{-1,1\}^n$ is the vector of labels. See Appendix~\ref{ap:logistic} for more details.
 
 \vspace{-4mm}
\subsection{Lasso and $\ell_1$-norm}
\label{sec:lasso}
\vspace{-2mm}
Following~\citet{woodworth2020kernel,vaskevicius2019implicit,pesme2021implicit}, we consider ``diagonal networks''. All of the three mentioned works  achieve implicit sparsity regularization  under sufficiently small initialization of the model parameter. Here we show that noise injection in the model also induces an $\ell_1$-regularization in the problem. The main advantage we have over the discussed work is that our approach is applicable to more complex models as we show below.

We consider $w = (w_1,w_2) \in \rb^{2d}$, and $\Phi(w) = X ( w_1 \circ w_1 - w_2 \circ w_2)$ for $X \in \rb^{n \times d}$, $L(\varphi) = \frac{1}{2n} \| y - \varphi \|_2^2$, with $m=2d$. When the model is overparametrized, that is $X$ has rank $n$, we can apply Theorem~\ref{theo:over}, and we then get an equivalent risk: 
\vspace{-2mm}
\begin{align} \label{eq:def_Lasso}
\textstyle R_\sigma^{({\rm eff})} (w)  =\frac{1}{2n} \| y -  X ( w_1 \circ w_1 - w_2   \circ w_2)\|_2^2 \ \notag \\
+ 2 \sigma^2  \diag(X^\top X/ n)^\top ( w_1 \circ w_1 + w_2 \circ w_2),
\end{align}
which is exactly the Lasso once considering $\beta = w_1 \circ w_1 - w_2   \circ w_2$, that is, minimizing
\begin{equation}
\label{eq:def_LassoB}
  \textstyle  \frac{1}{2n} \| y -  X \beta \|_2^2 \ + 2 \sigma^2  \diag(X^\top X/ n)^\top | \beta| ,
\end{equation}
where $\beta = w_1 \circ w_1 - w_2 \circ w_2$ 
(an explicit derivation of Eq.~\eqref{eq:def_Lasso} is given in  App.~\ref{ap:formula_derivation}).
However, there exist many efficient algorithms  dealing with $\ell_1$-regularization \citep[see][and references therein]{bach2012optimization},  and one should probably not use our reduction to optimize Eq.~\eqref{eq:def_LassoB}.

\vspace{-2mm}
\subsection{Nuclear norm (linear networks)}
\label{sec:linear_networks}


Following~\citet{baldi1995learning,arora2018convergence,saxe2019mathematical,gidel2019implicit}, we consider $w = (W_1,W_2)$, with $W_1 \in \rb^{ d_1 \times d_0}$ and $W_2 \in \rb^{d_2  \times d_1 }$, and $\Phi(w) =   W_2 W_1 X^\top $ with input data $X \in \rb^{n \times d_0}$. We consider the square loss for simplicity, that is, $L(\varphi) = \frac{1}{2n} \| Y^\top - \varphi\|_F^2$ for a response $Y \in \rb^{n \times d_2}$ and $\varphi \in \rb^{ d_2 \times n}$.\\
We can then apply Thm.~\ref{theo:over}, where the Hessian has no diagonal term, and we get (see detailed computations in App.~\ref{ap:formula_derivation}): 
\begin{align} \label{eq:R_sigma_nuclear_norm}
  \textstyle  R_\sigma^{({\rm eff})} (W_1,W_2)   =  \frac{1}{2n} \| Y^\top -  W_2 W_1 X^\top  \|_F^2 \notag \\
  + \frac{\sigma^2}{2n} \big[d_2  \| W_1X^\top \|_F^2 + \|W_2 \|_F^2 \cdot \| X\|_F^2 \big].
\end{align}
Given the matrix $M = W_2 W_1 \in \rb^{ d_2 \times d_0}$, we can optimize $\frac{1}{2} \big[d_2  \| W_1X^\top \|_F^2 + \|W_2 \|_F^2 \cdot \| X\|_F^2 \big]$ with respect to compatible matrices $W_1$ and $W_2$, leading to the penalty $\sqrt{d_2} \| X\|_F \cdot \|   M X^\top \|_\ast$, where $\| \cdot\|_\ast$ is the nuclear norm (sum of singular values), which favors low-rank matrices. Thus, minimizing $R_\sigma^{({\rm eff})} $ above is equivalent to 
 minimizing
\begin{equation}
 \textstyle   \frac{1}{2n} \| Y^\top -   MX^\top  \|_F^2 + \frac{\sigma^2}{n} \sqrt{d_2} \| X\|_F \cdot \|   M X^\top \|_\ast.
\end{equation}
We  obtain a nuclear norm penalty, thus providing an explicit regularizer in situations where the classical techniques for diagonal networks do not extend. See an empirical illustration in Section~\ref{sec:exp}.

We would also like to mention here a conjecture from \citet{gunasekar2017implicit}, which claims that with small enough step sizes and initialization close enough to the origin, gradient descent on a full dimensional factorization converges to the minimum nuclear norm solution, which we obtain explicitly using our noise injection approach without making those assumptions. Like for the Lasso, we are not advocating for gradient descent to be a particularly good algorithm for solving the problem above, although a detailed analysis is worthwhile performing as future work.


\subsection{Extensions}
\vspace{-2mm}
Other models could be considered as well, with relationships to other explicit regularizers.

\vspace{-1mm}

\textbf{Group Lasso.} It is traditional to recover the group Lasso as a special case of nuclear norm minimization \citep[see, e.g.,][]{bach2008consistency}. This is detailed in Appendix~\ref{ap:formula_derivation}, and can also be applied to the Lasso with a different formulation from Section~\ref{sec:lasso} (and no need for overparametrization).

\vspace{-1mm}
\textbf{Optimizing over the PSD cone.} By considering $w \in \rb^{d \times r}$, with $r \geq  d$ and $\Phi(w) = ww^\top \in \rb^{ d \times d}$, then for a cost function $L$ over symmetric matrices, we would get a trace penalty over positive semi-definite matrices.

\vspace{-1mm}

\textbf{Beyond quadratic models.}
Thus far, we have considered only functions $\Phi$ that were quadratic. This can be extended to higher order polynomials, like done for linear networks below, but we could imagine applications to tensor decomposition beyond the matrix decomposition above \citep{kolda2001orthogonal,kolda2009tensor,sidiropoulos2017tensor}. 

\section{One-hidden-layer neural networks with infinite widths}
\label{sec:neural-net}
\vspace{-2mm}
So far we have discussed the effect of noise injection on simple machine learning models. In this section, we will focus our attention on more complex models like neural networks with potentially large widths (where the limit of large widths is taken \emph{before} the limit of small $\sigma$). In this section, we start with a single hidden layer, and extend to deeper networks in Section~\ref{sec:neural-net-deep}.\\
We first start with linear networks, before considering ReLU activation functions in Sec.~\ref{sec:relu-single}. For simplicity, we only consider the (multivariate) square loss in our analysis.
\vspace{-4mm}
\subsection{Exploding variance for linear networks}
\label{sec:explosion}
\vspace{-2mm}

As already shown in Section~\ref{sec:linear_networks}, we consider
$W_1 \in \rb^{ d_1 \times d_0}$ and $W_2 \in \rb^{d_2  \times d_1 }$, and $\Phi(W_1,W_2) =   W_2 W_1 X^\top $ with input data $X \in \rb^{n \times d_0}$. We consider $L(\varphi) = \frac{1}{2n} \| Y^\top - \varphi\|_F^2$ for $Y \in \rb^{n \times d_2}$, the matrix of labels (we allow here for multi-dimensional outputs).

We here consider the overparametrized limit $d_1 \to +\infty$ using initialization with random~\footnote{The same analysis holds of course for Glorot initialization~\cite{glorot2010understanding}, with similar results.} weights of order $(W_1)_{ij} \sim \frac{1}{\sqrt{d_1 d_0}}$ for all $i,j$, that is $\|W_1\|_F^2$ not exploding with $d_1$, and $(W_2)_{ij} \sim \frac{1}{\sqrt{d_2 d_1}}$, that is $\|W_2\|_F^2$ not exploding with $d_1$.

We have, with Gaussian perturbations $E_1$ and $E_2$, an explicit exact expansion for linear networks:
\begin{align*}
&\Phi(w+ \sigma \varepsilon)\\
&=   (W_{2}+\sigma E_2)(W_1+\sigma E_1) X^\top \\
  & =  W_2 W_1X^\top + \sigma (  W_2E_1 + E_2W_1 )X^\top + \sigma^2  E_2 E_1 X^\top. 
\end{align*}
Taking expectations and using that $E_1,E_2$ have zero mean and are independent, and such that, $\E [ E_i M E_i^\top ] = \tr(M) \idm $ for $i=1,2$, and $M$ any symmetric matrix of compatible size, we can get $\E \big[ \Phi(w+ \sigma \varepsilon)  \big]
 = \Phi(w)$, and:
 \vspace{-2mm}
\begin{align*}\!
\E \big[ \|  \Phi(w+ \sigma \varepsilon)\|_F^2 \big]
   \! =\!  \| \Phi(w)\|_F^2 + \sigma^2   \big[\| W_2 \|_F^2 \| X\|_F^2 \\ + d_2
 \| W_1 X^\top\|_F^2   \big]  + \sigma^4    
 d_2 d_1 \| X\|_F^2
 .
\end{align*}
We can now compute $R_\sigma$ as:
 \vspace{-2mm}
\begin{align*} \textstyle
\!R_\sigma(W_1,W_2)  =  R(W_1,W_2) 
+ \frac{\sigma^2}{2n} \big[\| W_2\|_F^2 \| X\|_F^2 \\ + d_2 
 \| X W_1^\top \|_F^2  \big]  + \frac{\sigma^4}{2n}  
 d_1 d_2 \| X\|_F^2
.
\end{align*}
We recover the expression from Section~\ref{sec:linear_networks}, but we have an extra term $\frac{\sigma^4}{2n} d_2 d_1 \| X\|_F^2$, which is of superior order in $\sigma$, but problematic  when $d_1 \to +\infty$.

Note that $ \frac{\sigma^2}{2n}  \tr ( W_2^\top W_2) \tr ( X^\top X )$ scales as $\frac{\sigma^2}{n} \| X\|_F^2 $, while the term $\frac{\sigma^2}{n} d_2 
 \tr (   W_1 X^\top X W_1^\top ) $ scales as $d_2 \frac{\sigma^2}{n} \| X\|_F^2$, with thus no explosion in $d_1$.
 However, the term 
 $\frac{\sigma^4}{2n} d_2 d_1 \| X\|_F^2$ explodes when $d_1$ goes to infinity. \textit{This is the exploding variance problem of perturbing all layers simultaneously} --- see discussion in the experiment section as well as Figures~\ref{fig:hidden_toy} and \ref{fig:FMNIST_summary}. We provide a fix by perturbing one layer at a time, which we now present. 
 
 \begin{figure}
     \centering
     \includegraphics[width=\linewidth]{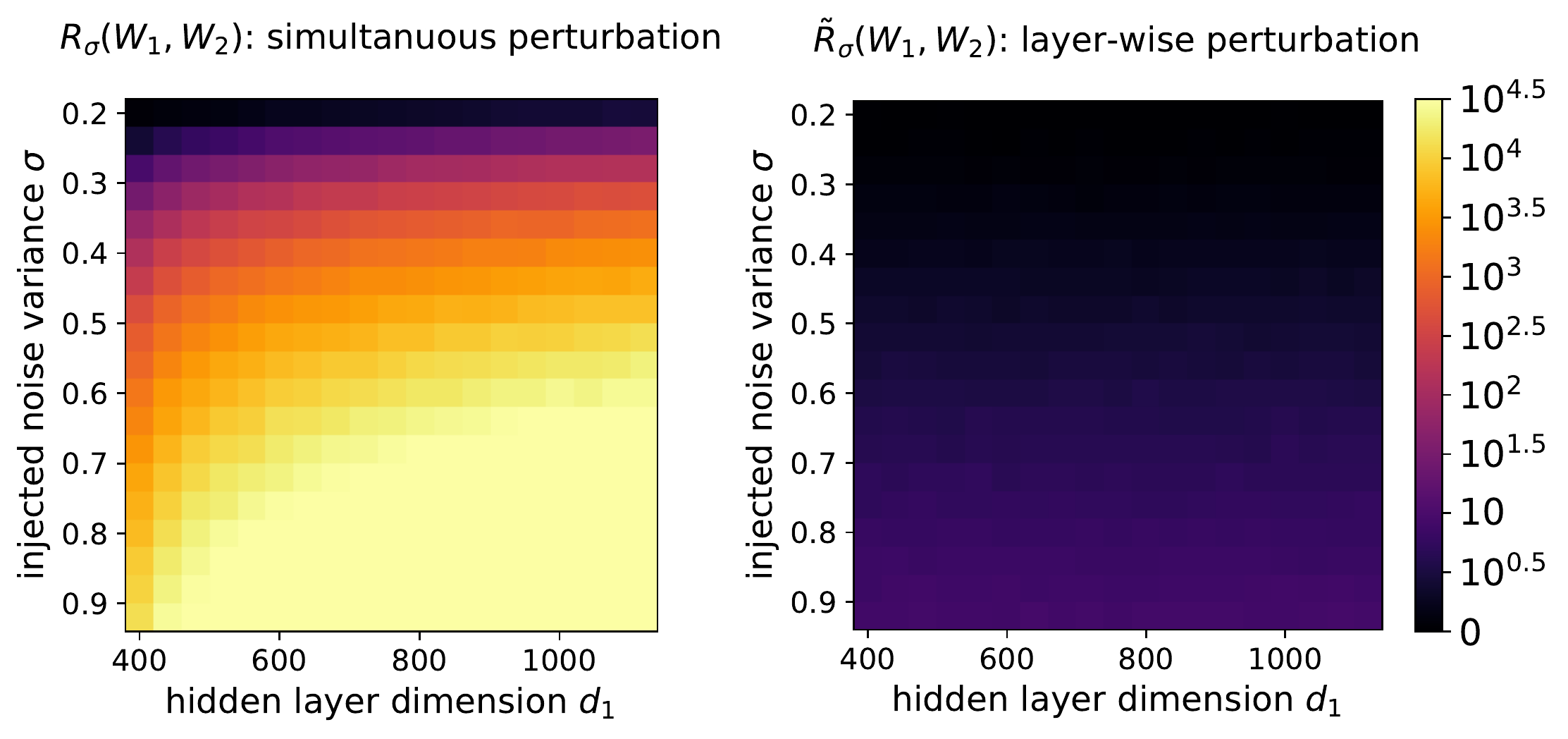}
     \vspace{-6mm}
     \caption{\small Numerical illustration of the analysis in Sec.~\ref{sec:explosion}\&\ref{sec:relu-single} on a linear network with 1 hidden layer of dimension $d_1$~(ReLU case and deep networks studied in the coming sections). If noise with standard deviation $\sigma$ is added to all weights~\textbf{(left)}, then the regularized loss $R_{\sigma}$ explodes as $d_1\to\infty$ due to the variance term $\sigma^4 d_2 d_1 \|X\|^4_F$. Instead, perturbing $W_1$ and $W_2$ in alternation \textbf{(right)} with standard deviation $\sqrt{2}\sigma$ provides mathematically the same expected regularization but avoids the variance term and therefore provides a much more reliable regularization, as clear also from the experimental section. Runs are repeated 100 times and shown is the average. }
     \label{fig:layer_vs_all}
     \vspace{-5mm}
 \end{figure}
 
 \subsection{Layer-wise perturbations for linear networks}
 \label{sec:layer_wise_linear}
 \vspace{-2mm}
 We consider the selection of one layer $j \in \{1,2\}$ out of the two layers at random, and add $\sqrt{2} E_j$ to $W_j$ before computing the gradient. This corresponds to minimizing an objective function $\tilde{R}_\sigma(W_1,W_2)$, which is   an expectation (both with respect to $j$ and $E_j$) equal to  
\begin{align*}
&\tilde{R}_\sigma(W_1,W_2)
 =  \textstyle \frac{1}{2} \E \big[ R(W_1 + \sigma\sqrt{2} E_1, W_2 )\big] \\
 &\qquad \qquad \qquad \qquad + \frac{1}{2}\E \big[ R(W_1, W_2  + \sigma\sqrt{2}  E_2)\big]. 
 \end{align*}
Hence
\begin{align*}
&\tilde{R}_\sigma(W_1,W_2) =  \textstyle R(W_1,W_2) + \frac{1}{4n} \E \big[ \| \sigma\sqrt{2}  W_2 E_1 X^\top  \|_F^2 \big]\\
 & \qquad \qquad \qquad \qquad + \frac{1}{4n} \E \big[ \| \sigma\sqrt{2}  E_2 W_1 X^\top  \|_F^2 \big]
 \\
 & =   \textstyle R(W_1,W_2) 
+ \frac{\sigma^2}{2n} \big[\tr ( W_2^\top W_2) \tr ( X^\top X ) \\
& \qquad \qquad \qquad \qquad \qquad \qquad + d_2 
 \tr (   W_1 X^\top X W_1^\top )  \big],
\end{align*}
that is, the exploding variance term is not there anymore, and we arrive at the same nuclear norm interpretation as in Section~\ref{sec:linear_networks}.
This generalizes to non-linear activation functions as we show next.

\subsection{One-hidden layer with non-linear activations}
\label{sec:relu-single}
We now consider the same set-up as above, but with ReLU activations, that is,  $\Phi(w)
= W_2 ( W_1 X^\top )_+$, with $X \in \rb^{n \times d_0}$, and $L(\varphi) = \frac{1}{2n} \| Y^\top - \varphi\|_F^2$. We consider the square loss for simplicity, but this applies more generally.

We also consider  weights $W_1$ and $W_2$ initialized with respective scales $1/\sqrt{d_1 d_0}$
and $1/\sqrt{d_2 d_1}$. After rescaling, it corresponds exactly to the  mean-field limit that allows representation learning~\citep{chizat2018global,mei2018mean,wojtowytsch2020convergence,sirignano2022mean}, for which the scaling s.t. $\|W_2\|_F^2$ and $\|W_1\|_F^2$ do not explode is preserved throughout optimization.

Here, we consider asymptotics explicitly with respect  $d_1$ (going to infinity), and then with respect to $\sigma$ (going to zero). We show in App.~\ref{ap:sec_3_neural_net} that the same variance explosion as in Sec.~\ref{sec:explosion}, when perturbing all layers simultaneously, still occurs in the limit of infinite widths and non-zero $\sigma$. We thus only focus here on the layer-wise perturbation. 

Locally, we obtain
$
\Phi(W_1 + \sigma\sqrt{2} E_1,W_2 ) = \Phi(W_1,W_2) + \sigma\sqrt{2}  W_2 \big[( W_1 X^\top)_+^0 \circ E_1 X^\top \big]   $
and $
\Phi(W_1,W_2 + \sigma\sqrt{2} E_2)=  \Phi(W_1,W_2) + \sigma\sqrt{2}   E_2 ( W_1 X^\top)_+$; see the Appendix~\ref{ap:sec_3_neural_net} for a formal derivation. 
We can therefore compute $\tilde{R}_\sigma$ as: 
\begin{align*}
\tilde{R}_\sigma(W_1,W_2)
=  R(W_1,W_2) + \frac{\sigma^2}{2n} \| (W_1 X^\top)_+ \|_F^2 \\ + \frac{\sigma^2}{2n} \sum_{j=1}^{d_1} \sum_{i=1}^n \| (W_2)_{\cdot j}\|_2^2 \times 
| (( W_1 X^\top)_+^0)_{ji} |^2.
\end{align*}
Note that here our regularizer is 2-homogeneous, like considered by~\citet{chizat2020implicit}, but that the penalty is not any more separable in $W_1$ and $W_2$. Like done by~\citet{chizat2020implicit}, we can take the number $d_1$ of hidden neurons to infinity and study the associated function space norm, which has similar adaptivity properties (see the Appendix~\ref{ap:sec_3_neural_net} for details).

\section{Deeper networks}
\label{sec:neural-net-deep}

In this section, we show that we still get the explicit effect of perturbing each layer selected randomly with standard deviation $\sigma \sqrt{M}$, where $M$ is the number of layers, and the extra $\sqrt{M}$ term is added to take into account that the expectation with respect to the choice of the layer adds a multiplicative factor $1/M$. The main purpose of this section is to show that the expression of $R_\sigma^{({\rm eff})}$ found for finite-dimensional models in Section~\ref{sec:finitedim}, is still valid, that is:
\begin{equation}
\label{eq:deepRs}
   \textstyle  R_\sigma^{({\rm eff})} (w) = R(w) +  \frac{\sigma^2}{2} \tr \big(
    \nabla^2  L(\Phi(w))  {\rm D}\Phi(w) {\rm D}\Phi(w)^\top
    \big).
\end{equation}
Crucially, note that with full perturbations, and with a number of neurons going to infinity, we would get a variance explosion (see discussion in Appendix~\ref{ap:sec_4_neural_net_deep}). This finding is also supported by our experimental findings in the next section.

In this section, to simplify the exposition and avoid technicalities, we only consider linear networks (see Appendix~\ref{ap:sec_4_neural_net_deep} for non-linear ones).

\begin{figure*}[ht]
\vspace{-2mm}
    \centering
    \includegraphics[height = 0.245\textwidth]{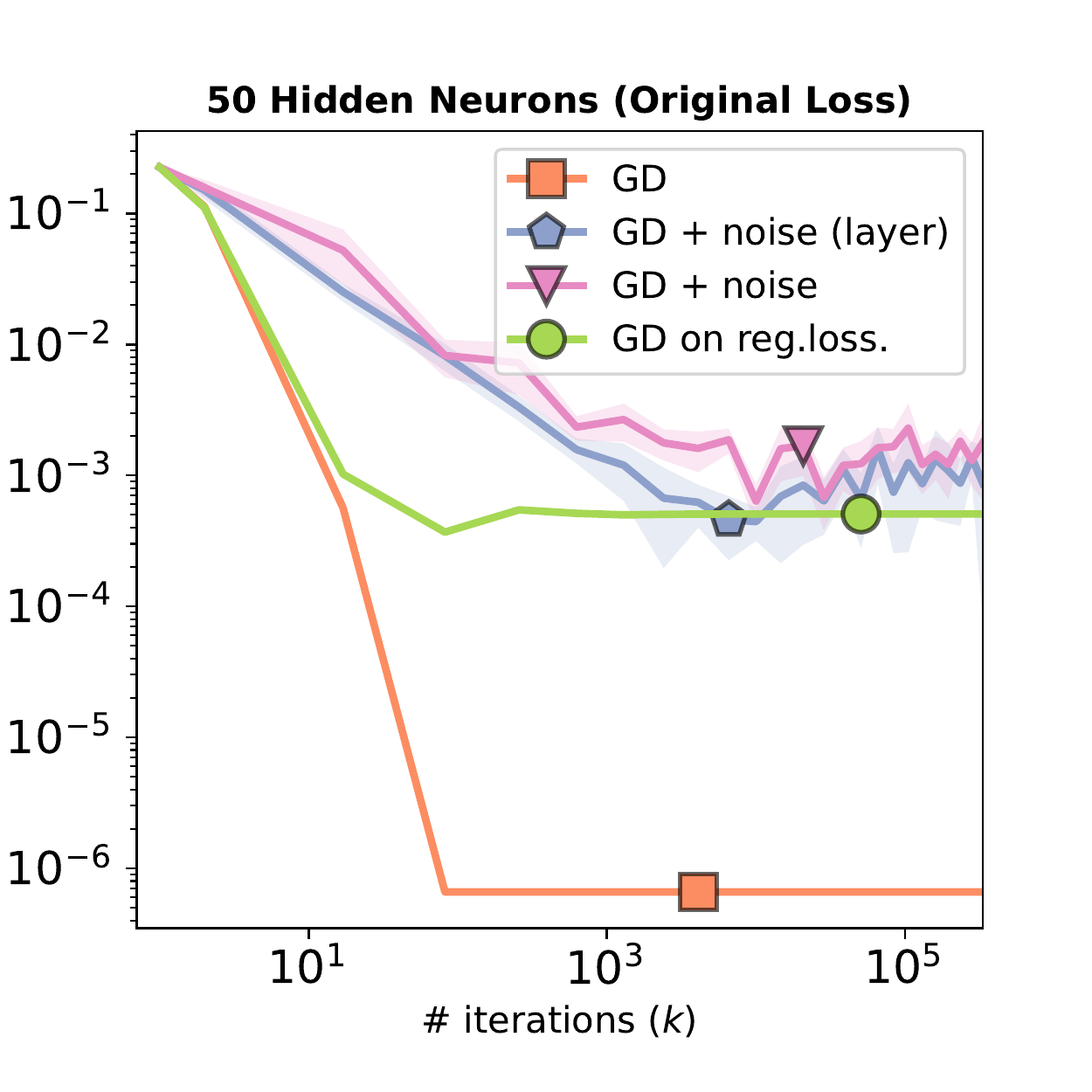}
    \includegraphics[height = 0.245\textwidth]{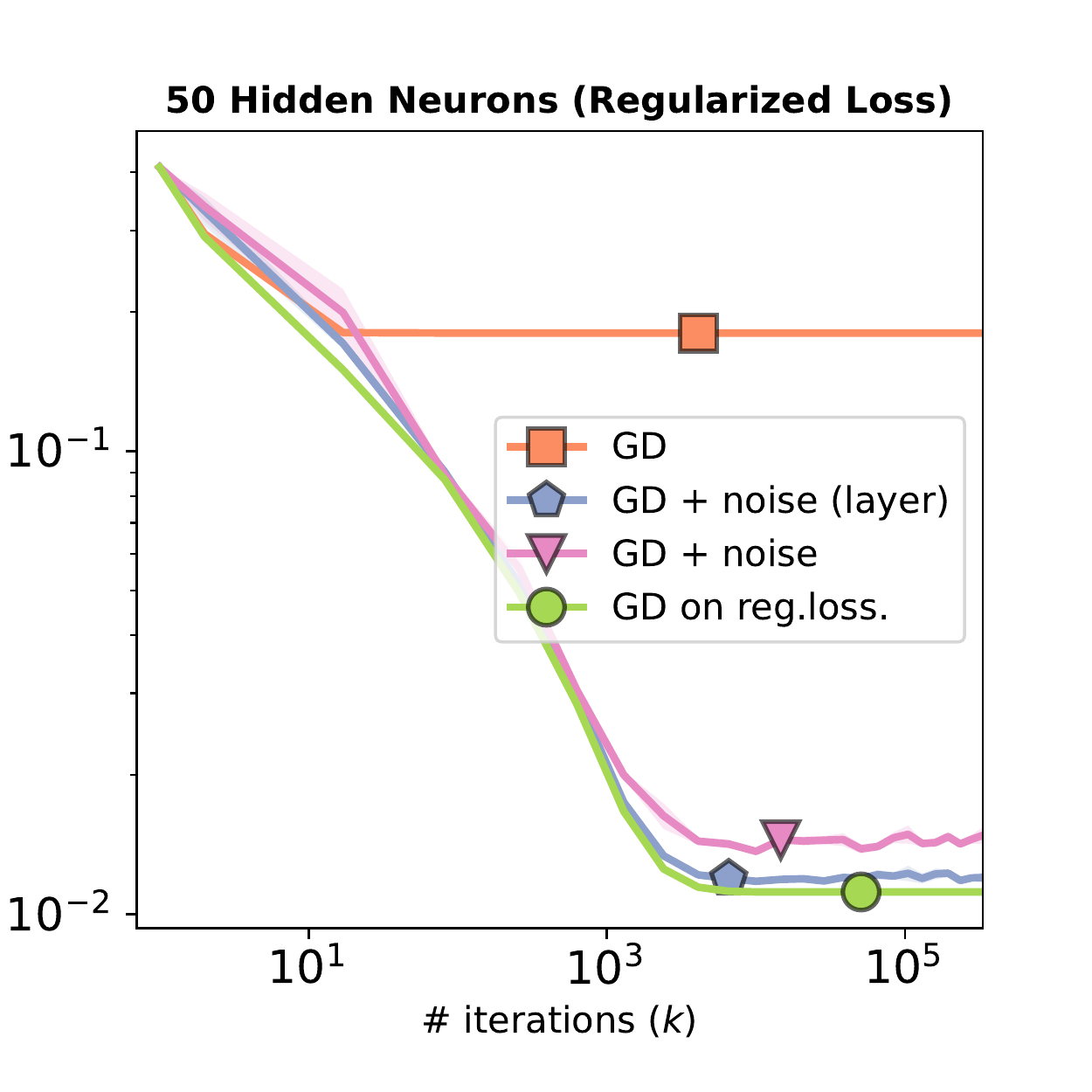}
    \includegraphics[height = 0.245\textwidth]{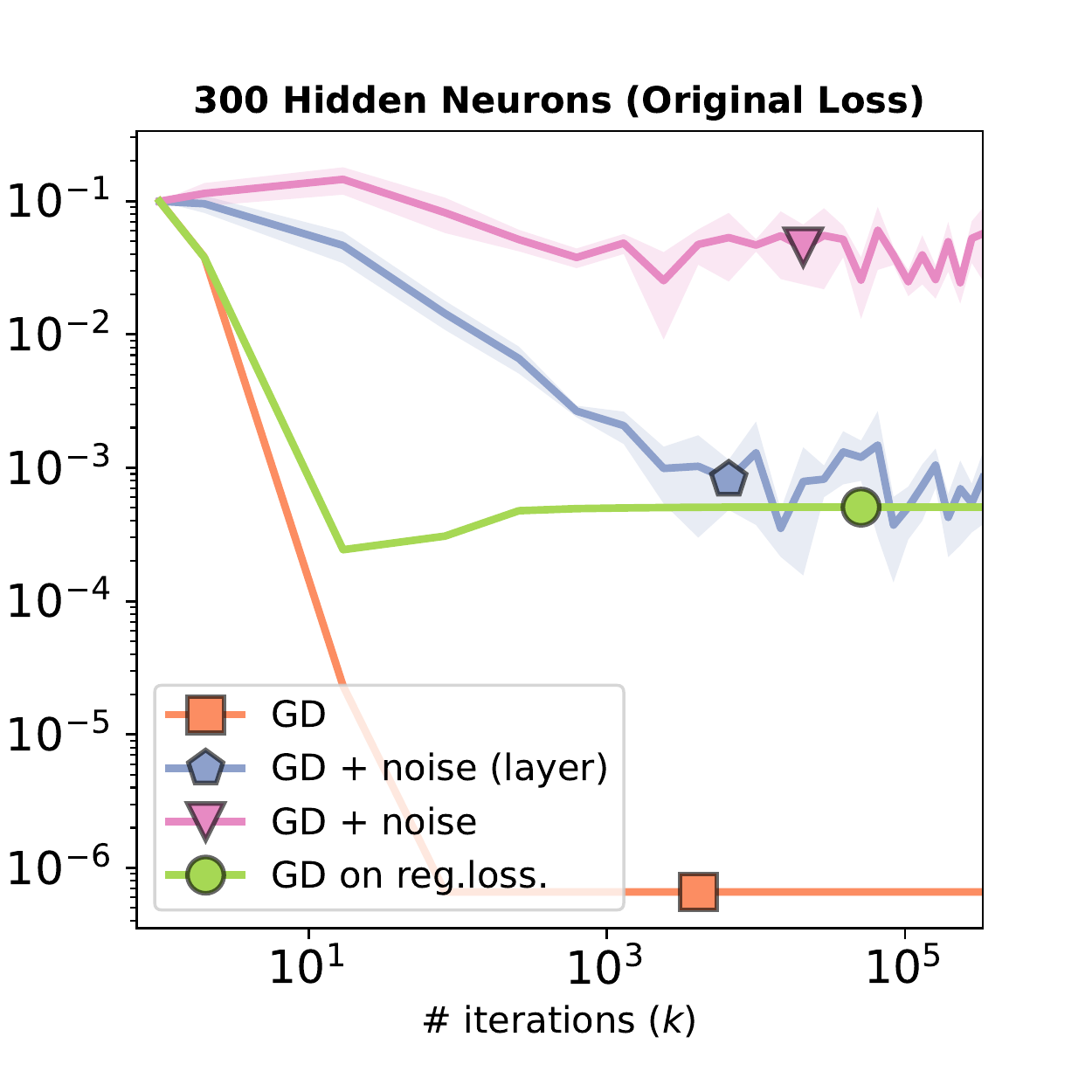}
    \includegraphics[height = 0.245\textwidth]{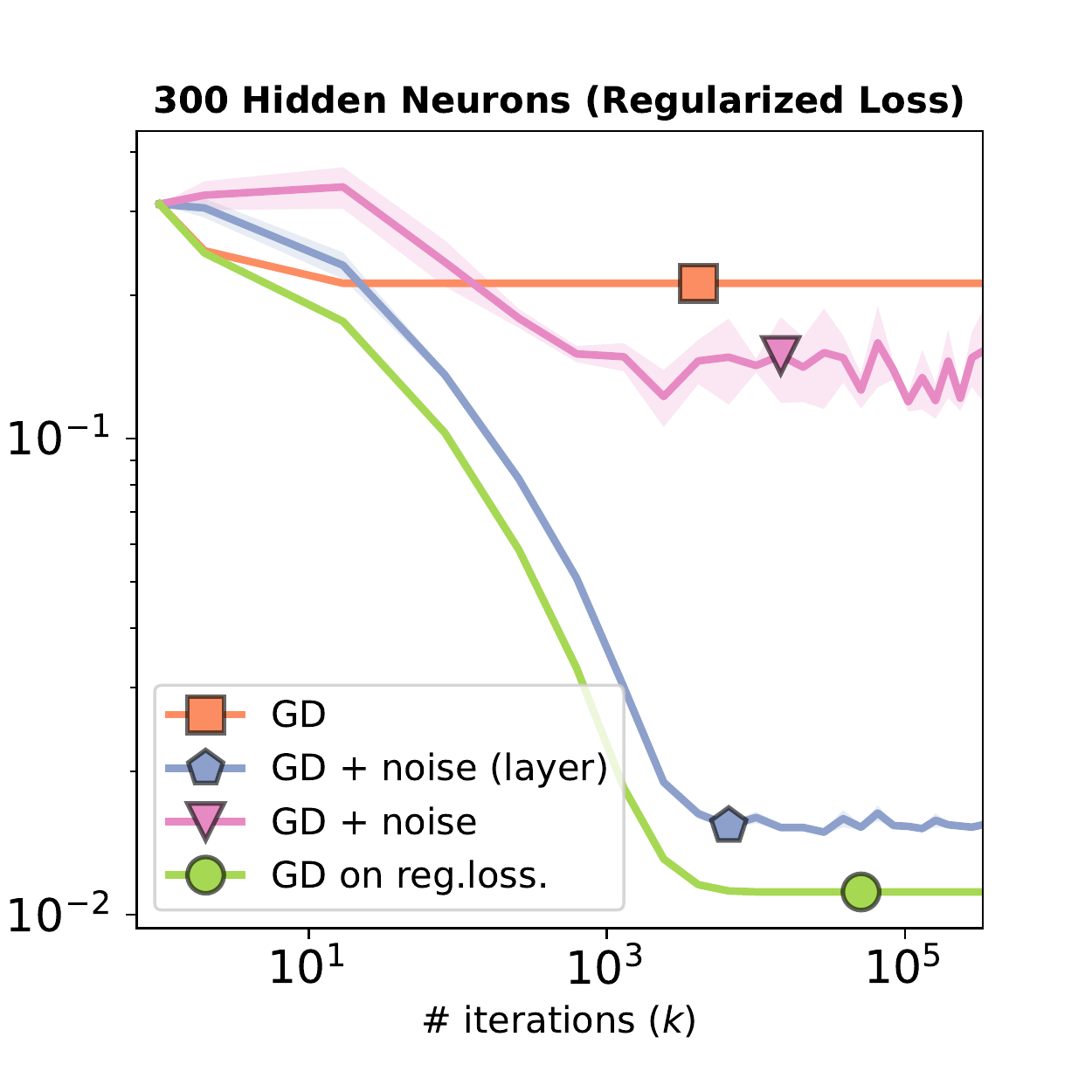}
    \vspace{-4mm}
    \caption{\small MLP with 1 hidden layer and linear activations on synthetic data~(see the Appendix for a full description and loss curves). For 50 hidden neurons (\textbf{left} panels) both injecting noise layer-wise and to all weights lead to minimization of the regularized loss. If we have 300 hidden neurons (\textbf{right} panels), then only injecting noise layer-wise takes us close to a minimizer for the regularized loss. For all methods the learning rate is set to $0.1$.}
    \vspace{-1mm}
    \label{fig:hidden_toy}
\end{figure*}

\textbf{Linear networks.}
 Consider a linear neural network with $w = (W_1,,\dots, W_M )$ with $W_i \in \mathbb{R}^{d_i \times d_{i-1}}$, and  $\Phi(w) = W_M W_{M-1} \cdots W_1 X^{\top} \in \rb^{d_M \times n}$ with the input data $X\in \mathbb{R}^{n\times d_0}$, with $L(\varphi) = \frac{1}{2n}\|Y^\top - \varphi \|_F^2$ for output data $Y \in \rb^{n \times d_M}$ and $\varphi \in \rb^{ d_M \times n}$. We now analyze the effect of independent layer-wise perturbations. 
We consider the selection of one layer $j$ out of the $M$ layers at random, and add a random Gaussian matrix $\sqrt{M} E_j$ to $W_j$. We then have an expectation (both with respect to $j$ and $E_j$), equal to
\vspace{-1mm}
\small
\begin{align*}
 &\tilde{R}_\sigma(W_1,\dots,W_M)\\
&=  \frac{1}{M} \sum_{j=1}^M
\E \big[ 
R(W_1,\dots,W_{j},W_j + \sqrt{M} \sigma E_j, W_{j+1}, \dots, W_M \big] \\
  &=
\frac{1}{M} \sum_{j=1}^M
\E \big[ L ( W_M W_{M-1} \cdots W_1 X^{\top} 
\\
&\qquad \qquad \qquad   +  \sqrt{M} \sigma W_M \cdots W_{j+1} E_j W_{j-1} \cdots W_1 X^\top ) \big] .
\end{align*}
We can then compute expectations and get
\begin{align*}
&\tilde{R}_\sigma(W_1,\dots,W_M)=
 \textstyle L ( \Phi(W_1,\dots,W_M))
\\
&\qquad \qquad  + \frac{\sigma^2}{2n}
\sum_{j=1}^M
 \big\| W_M \cdots W_{j+1} \big\|_F^2  \big\| W_{j-1} \cdots W_1 X^\top\big\|_F^2,
\end{align*}
\normalsize
which is exactly Eq.~\eqref{eq:deepRs}. We thus have a non-exploding explicit regularizer, based on the gradient of the prediction function \emph{with respect to the parameters}. We show in Section~\ref{sec:exp} that this extra regularization does improve test accuracy, and leave for future work the detailed theoretical study of this new regularizer.

\section{Experiments}
\label{sec:exp} 

The goal of this section is to provide experimental evidence\footnote{{\url{https://github.com/aorvieto/noise_injection_overparam}}} to back-up the results of the last sections. In particular, we compare gradient descent~(GD) with the perturbed variants studied in this paper: (1) noise injection at each weight in the model, for each iteration~(named ``\textit{GD+noise}''), and (2) layer-wise perturbations~(injection layer is sampled at random) at each iteration ~(named ``\textit{GD+noise (layer)}''). All perturbations are performed before the gradient computations. As seen in Section~\ref{sec:layer_wise_linear}, if the variance of layer-specific weight perturbations is $\sigma^2 M$, where $M$ is the number of layers, then the two noise injection methods minimize a similar regularized loss, in expectation. However, as the degree of overparametrization~(e.g., number of parameters) increases, the theory suggests that layer-wise perturbations are preferable since they overcome variance explosion~(see Sections~\ref{sec:explosion} and \ref{sec:layer_wise_linear}).

\textbf{Minimization of the regularized loss.} For a start, we consider a one hidden layer network with linear activations and either 50 or 300 hidden neurons on a randomly generated sparse\footnote{We consider 40 data points sampled from a Gaussian in 10 dimensions. The solution is sparse and is planted as the result of a linear prediction from the first two dimensions. } synthetic regression data set with inputs in $\mathbb{R}^{10}$ and outputs in $\R$. In Figure~\ref{fig:hidden_toy}, we show the dynamics of the original and regularized loss derived in this paper. As the theory predicts, minimization of the regularized loss in the strongly overparametrized setting is only achievable with layer-wise perturbations~(see Section~\ref{sec:layer_wise_linear}). As discussed, to keep the same explicit regularization, the variance of the noise in the layer-wise approach is doubled compared to the vanilla approach. This depth scaling is adopted for all further experiments.

\begin{figure}[ht]
    \centering
    {\scriptsize\textbf{Fashion MNIST - Wide Shallow MLP}}
    \includegraphics[width = 0.6\linewidth]{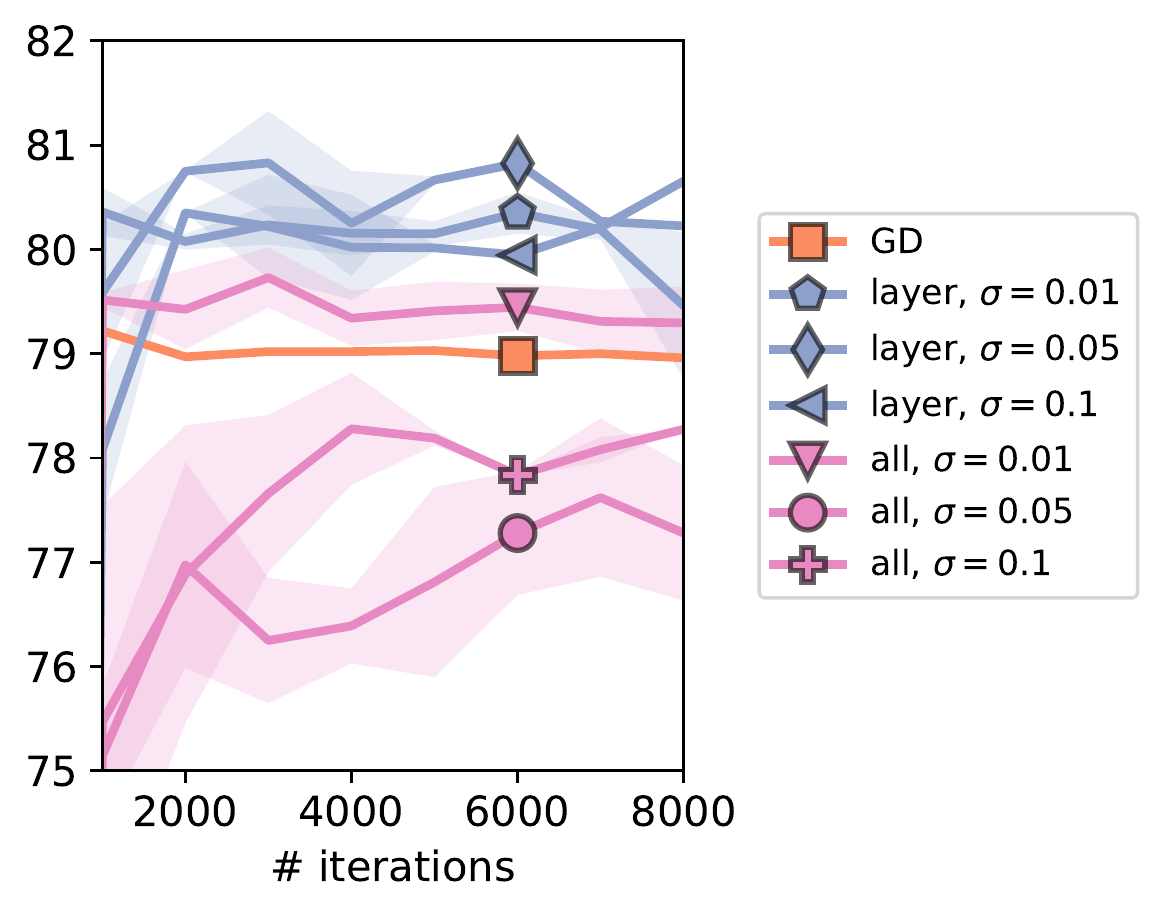}
    \vspace{-3mm}
    \caption{\small Test accuracy for an MLP on FMNIST~(2 HL with 5000 neurons). Comparison of perturbation effects.}
    \label{fig:tuning}
\end{figure}

\begin{figure}[ht]
\vspace{-5mm}
\hspace{-5mm}
    \centering
    \includegraphics[height = 0.24\textwidth]{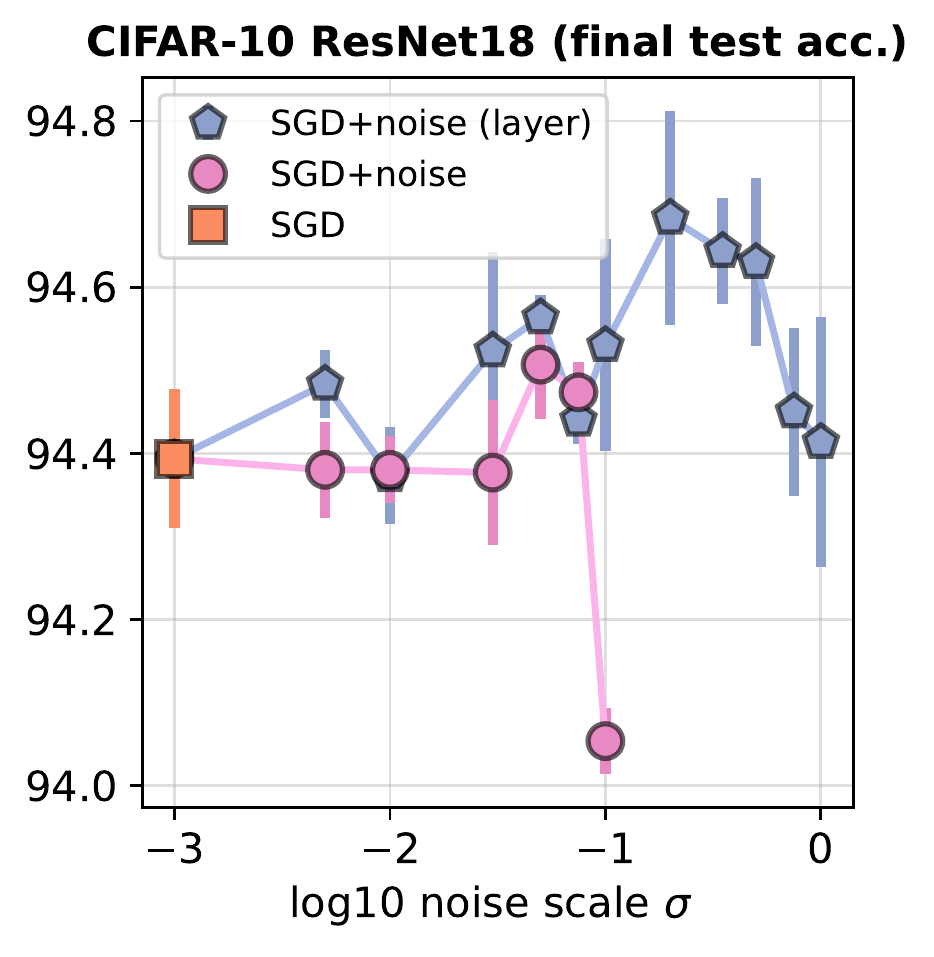}
    \includegraphics[height = 0.24\textwidth]{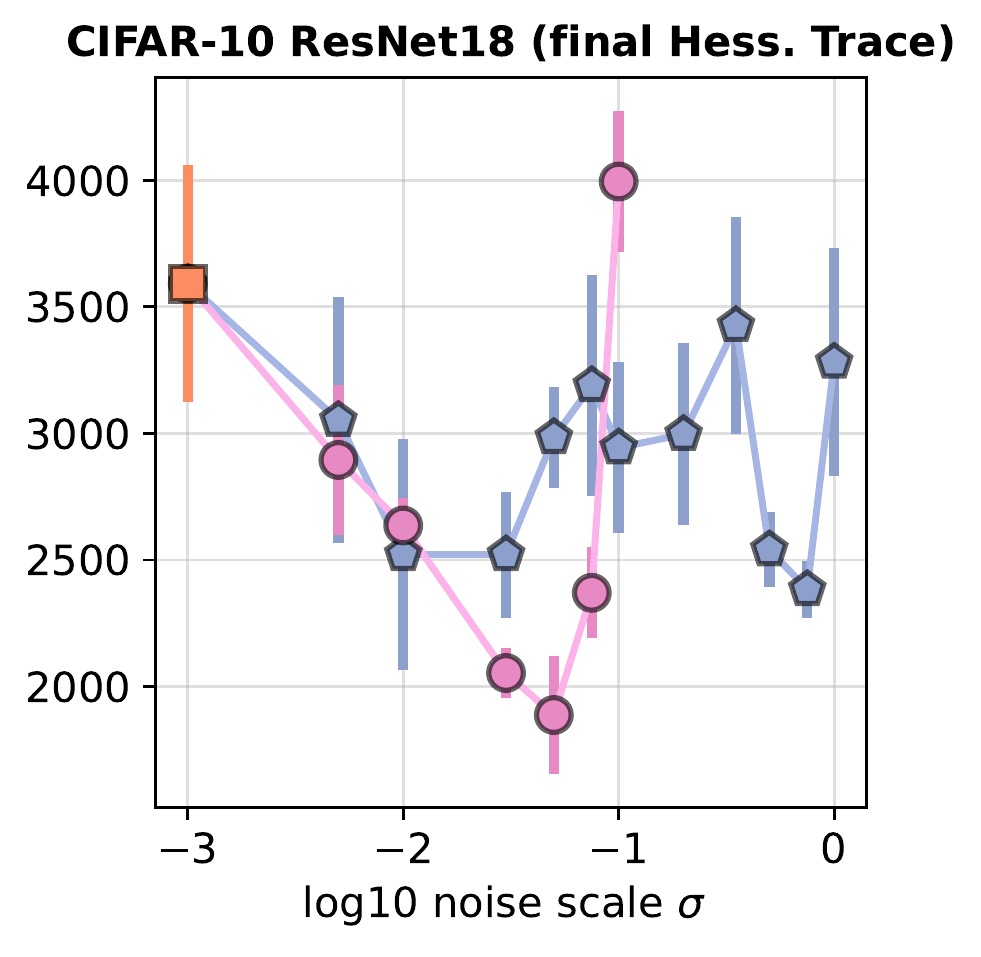}

    \vspace{-2mm}
    \caption{\small \textbf{Final} Test accuracy \textbf{(left)} and Hessian trace \textbf{(right)} for SGD on a CIFAR10 ResNet18 with batch normalization. Both plots show mean and standard error of the mean (3 runs). SGD is plotted at $\sigma=1e-3$ instead of $\sigma=0$ for better visualization. Layer-wise noise injection is able to boost performance by explicit regularization with a high $\sigma$. Instead, for such high $\sigma$, injecting noise simultaneously in all parameters results in instabilities. We remind that the total noise injection variance in normalized in the two methods, as detailed in this section.}
    \label{fig:CIFAR_big}
\end{figure}

\textbf{Effect of tuning.} The findings of the last paragraph are reported for one specific value of $\sigma$. We test the influence of $\sigma$ on a slightly more complex model: an MLP with 2 hidden layers~(5000 neurons each) and ReLU activations on Fashion MNIST~\citep{xiao2017fashion}~(classification). Given that the data is relatively easy to fit, we train on a subset of $1024$ data points --- to induce heavy overparametrization. We run full-batch gradient descent with learning rate $0.005$, and plot the test accuracy evolution~(computed using $10K$ data points) for different values of~$\sigma$. Figure~\ref{fig:tuning} shows that, for this wide model, layer-wise perturbation acts as an effective regularizer that is able to increase the test accuracy. This is in contrast to standard noise injection~(cf. Section~\ref{sec:explosion}). A similar result also holds true for CIFAR10 on ResNet18~(see Figure~\ref{fig:CIFAR_big}).

\begin{figure*}
    \centering
    \hspace{-4mm}
    \includegraphics[height = 0.20\textwidth]{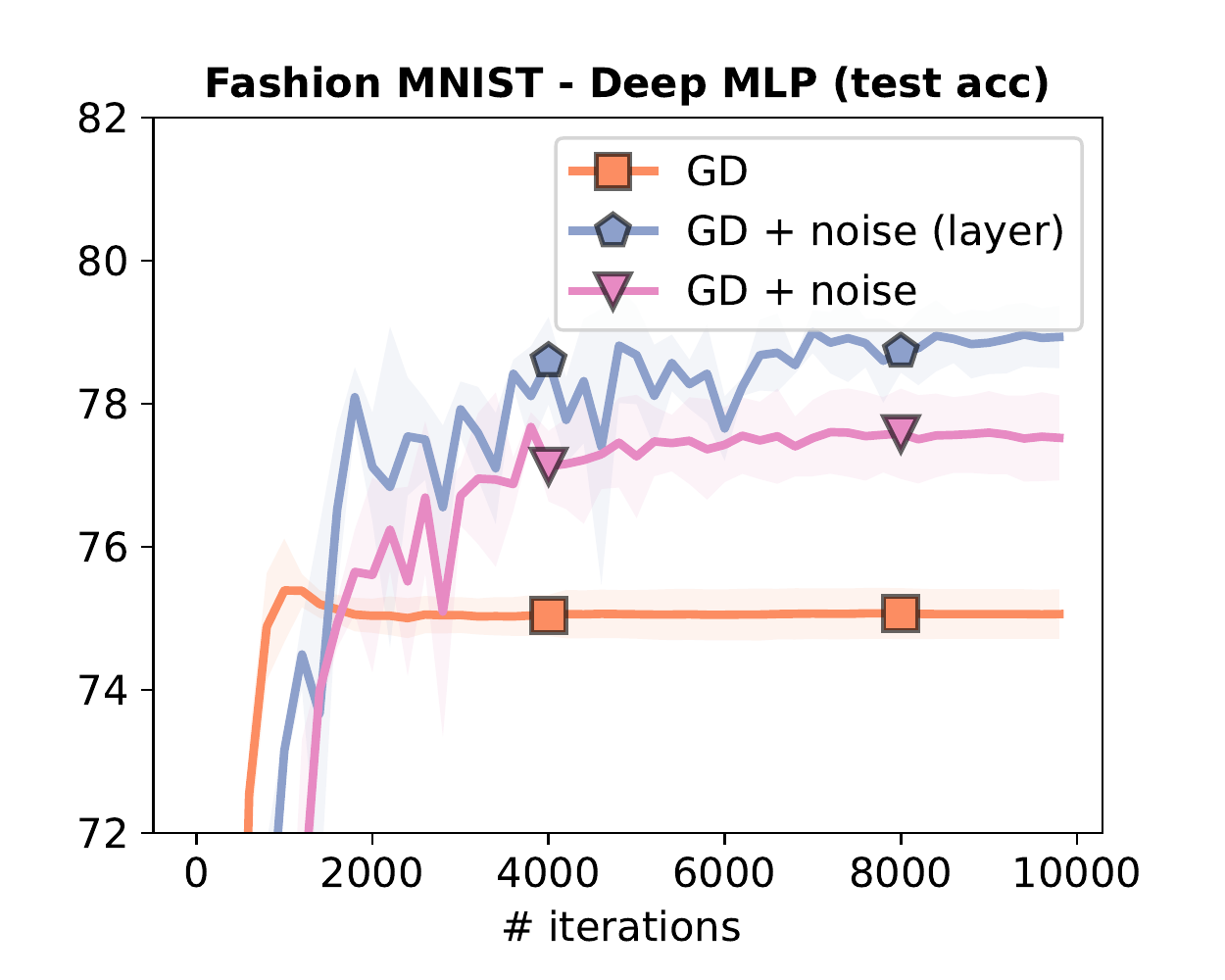}
    \includegraphics[height = 0.20\textwidth]{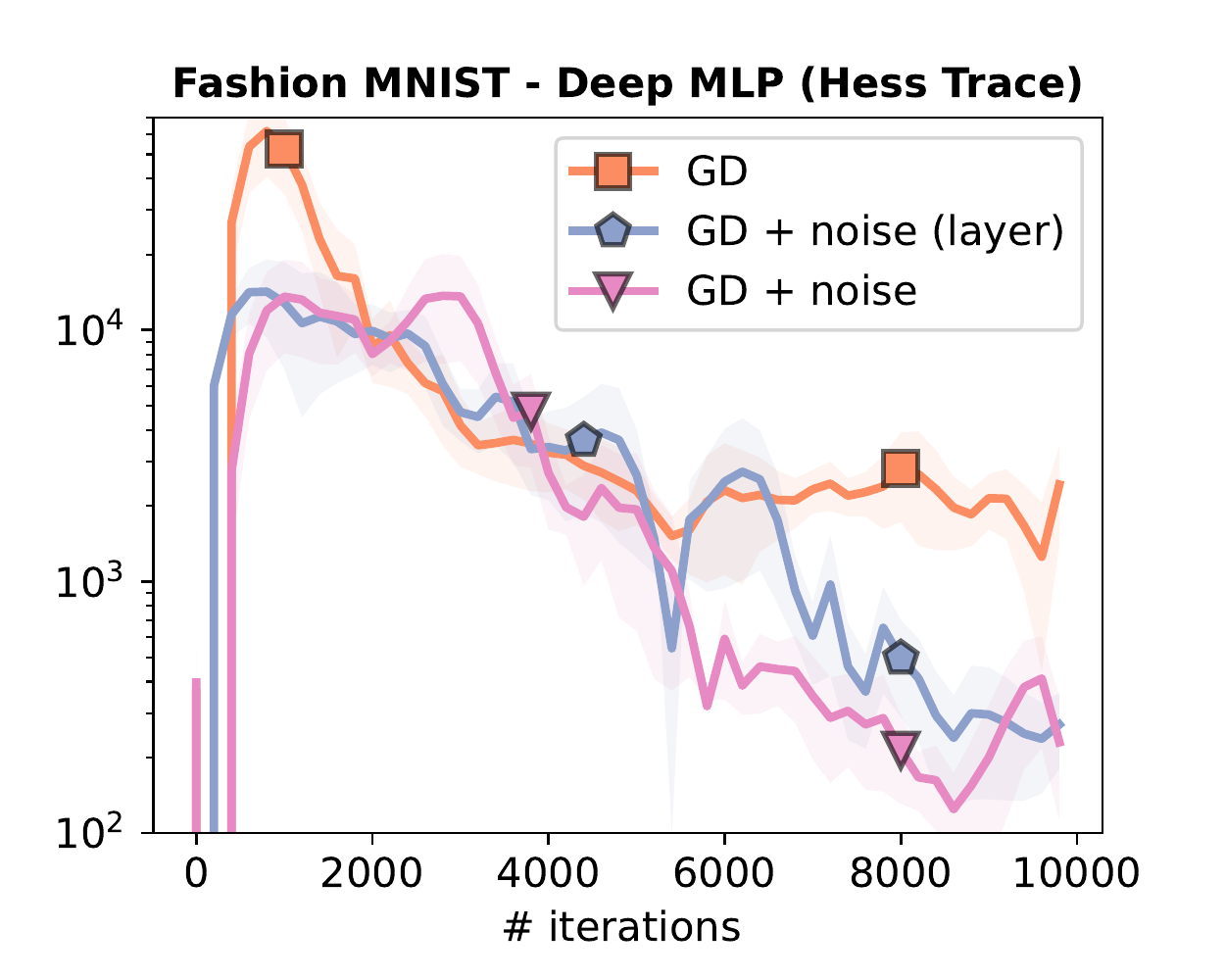}
    \includegraphics[height = 0.20\textwidth]{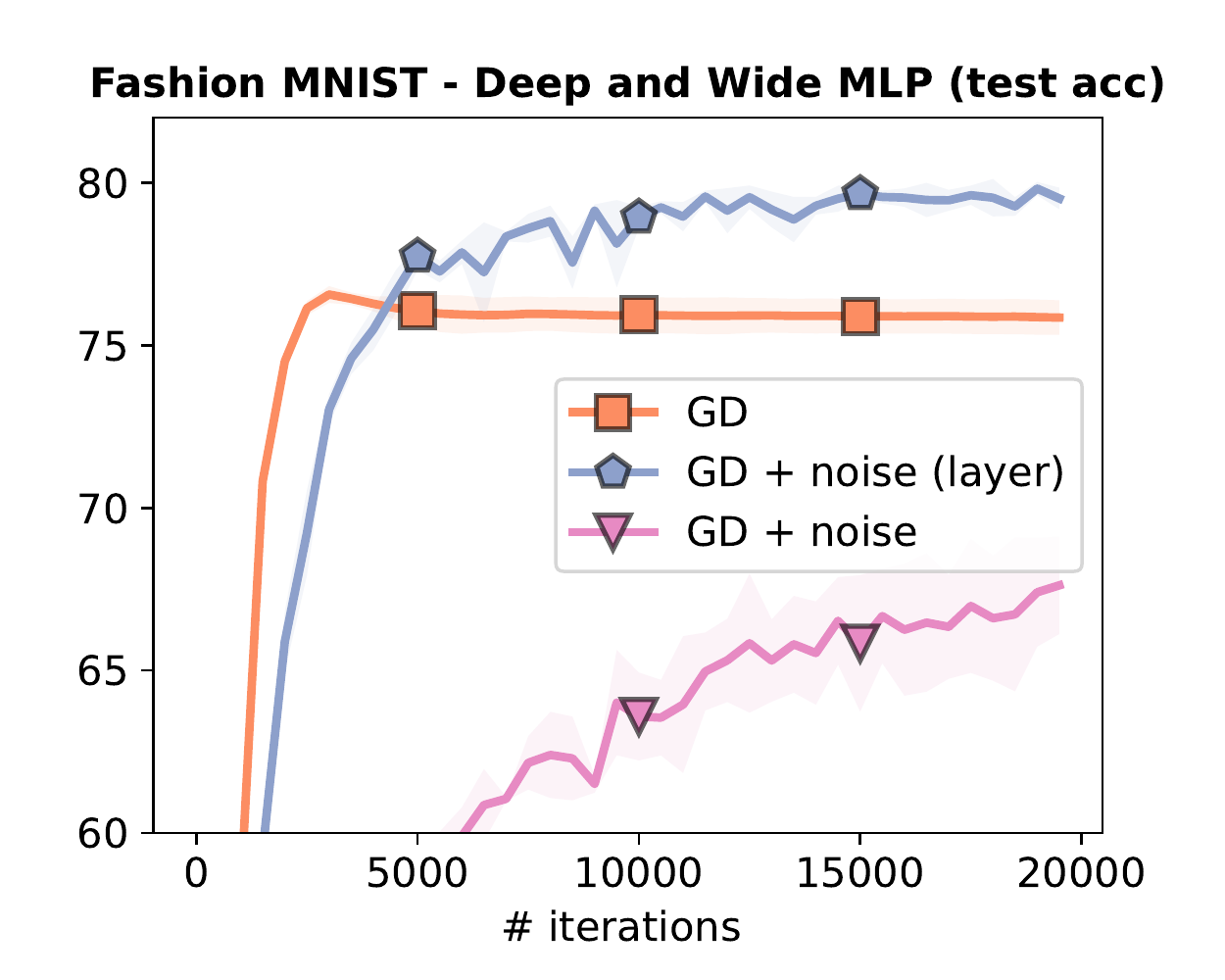}
    \includegraphics[height = 0.20\textwidth]{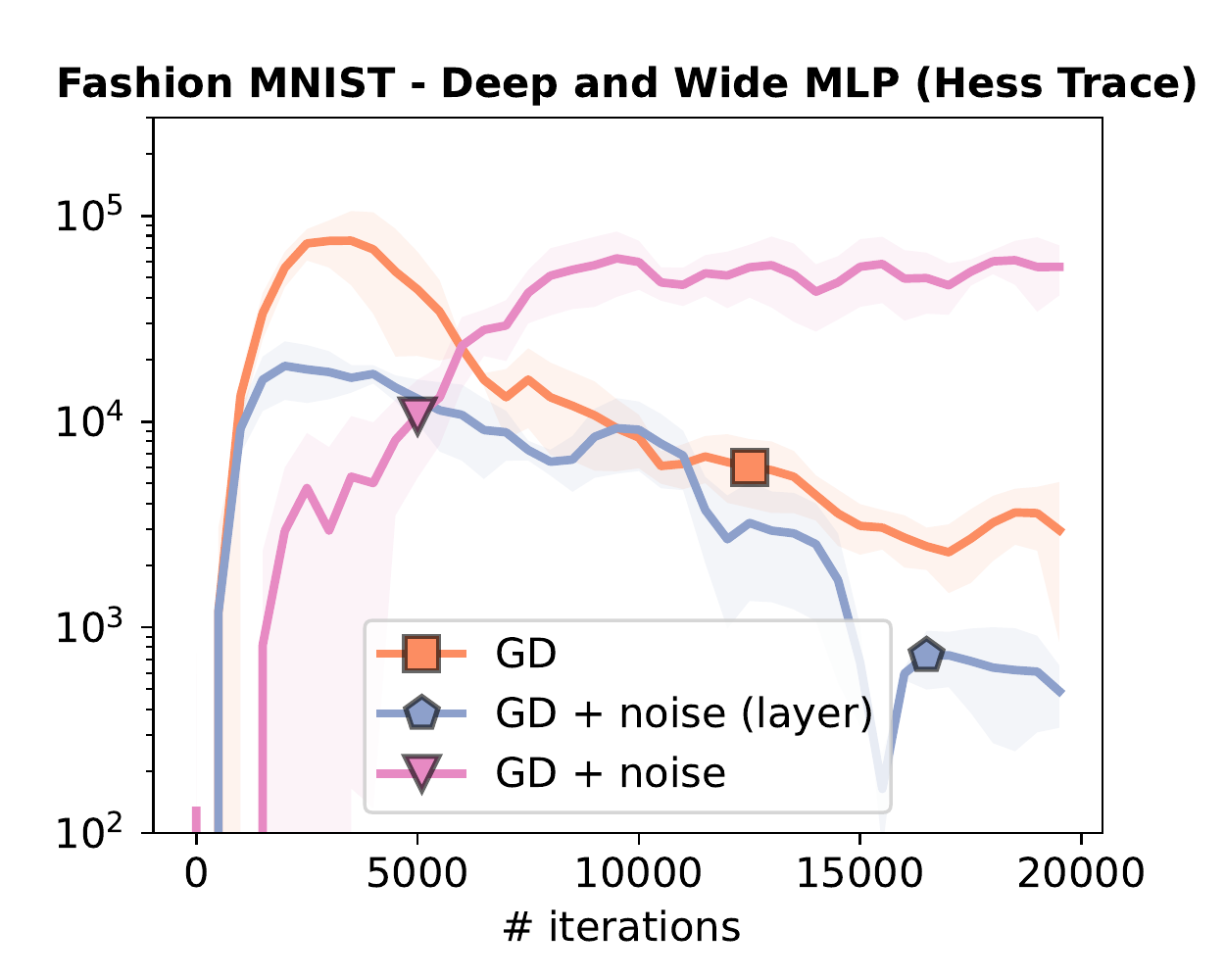}
    \vspace{-8mm}
    \caption{\small MLP with 5 hidden Layers and ReLU activations on a subset of Fashion MNIST. The hidden layers each have 1000 neurons (narrow, \textbf{left} panels) or 5000 neurons (wide, \textbf{right} panels). We run full-batch gradient descent. 
    The results show that injecting noise separately in each layer results in improved regularization and test accuracy, even in the overparametrized setting.}
    \label{fig:FMNIST_summary}
\end{figure*}

\textbf{Deep MLPs.} Next, we test how the findings carry over to deeper networks. In the same data setting as the last paragraph, we consider now a ReLU MLP with 5 hidden layers and either 1000 (narrow) or 5000 (wide) hidden neurons. In Figure~\ref{fig:FMNIST_summary} we test our methods~($\sigma = 0.05$) against full-batch GD, with a step size of $0.005$ in the narrow setting and $0.001$ in the wide setting, to account for the more complex landscape. As can be seen, again layer-wise perturbations yield the best result. This is also reflected in the size of the regularizer~(trace of Hessian), which is minimized by the best-performing method in terms of test error.

\begin{figure*}
\vspace{-2.5mm}
    \centering
    \hspace{-4mm}
    \includegraphics[height = 0.20\textwidth]{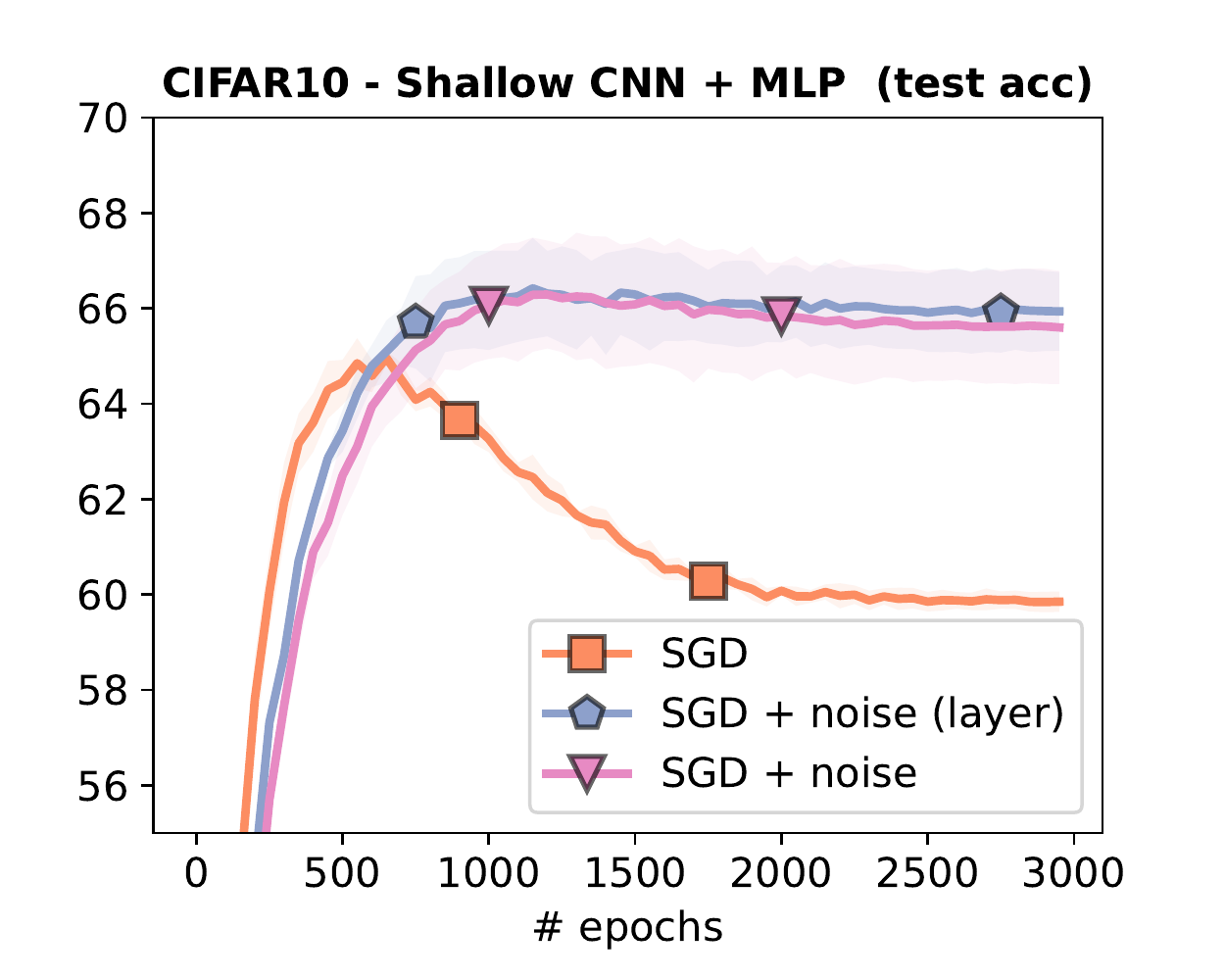}
    \includegraphics[height = 0.20\textwidth]{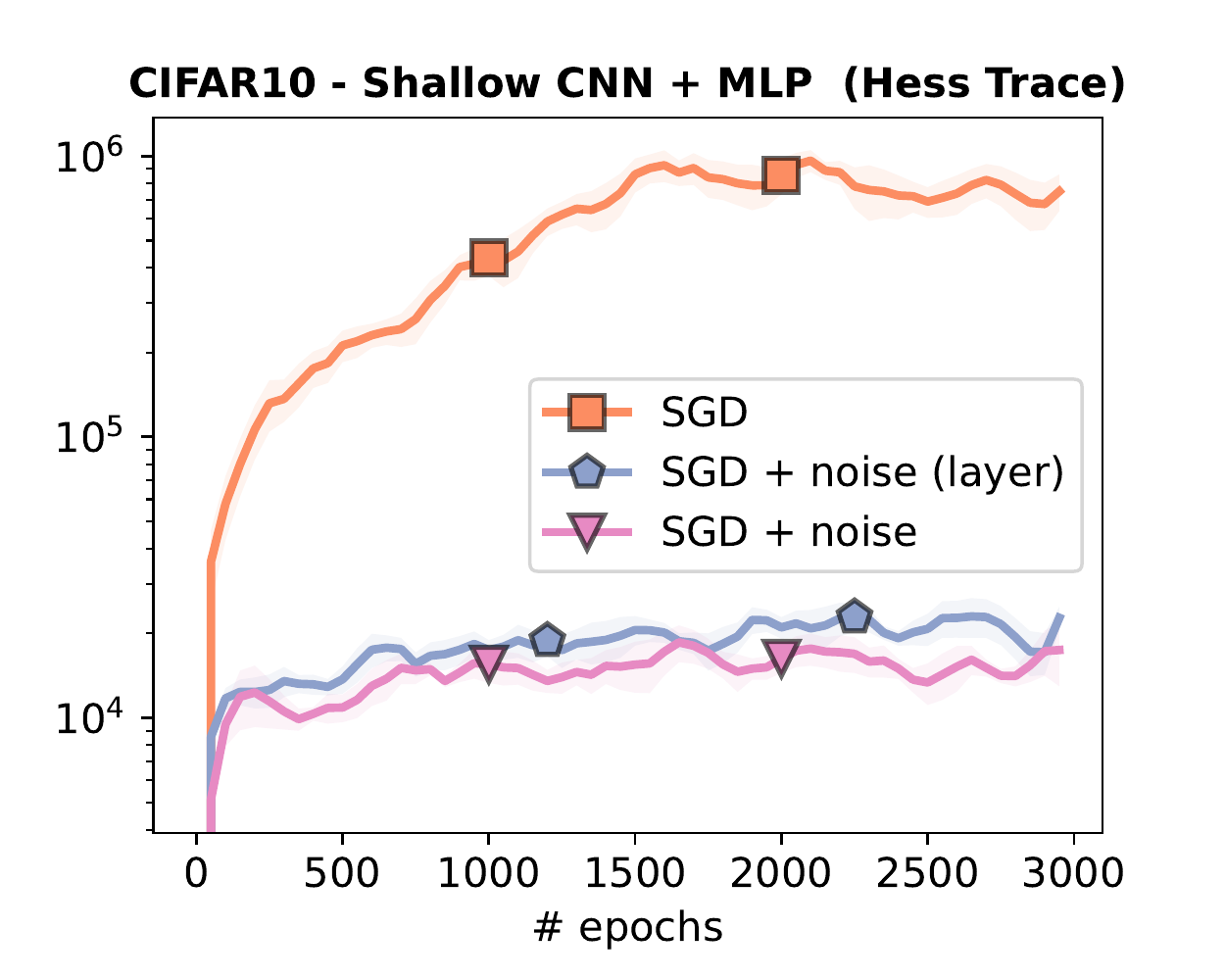}
    \includegraphics[height = 0.20\textwidth]{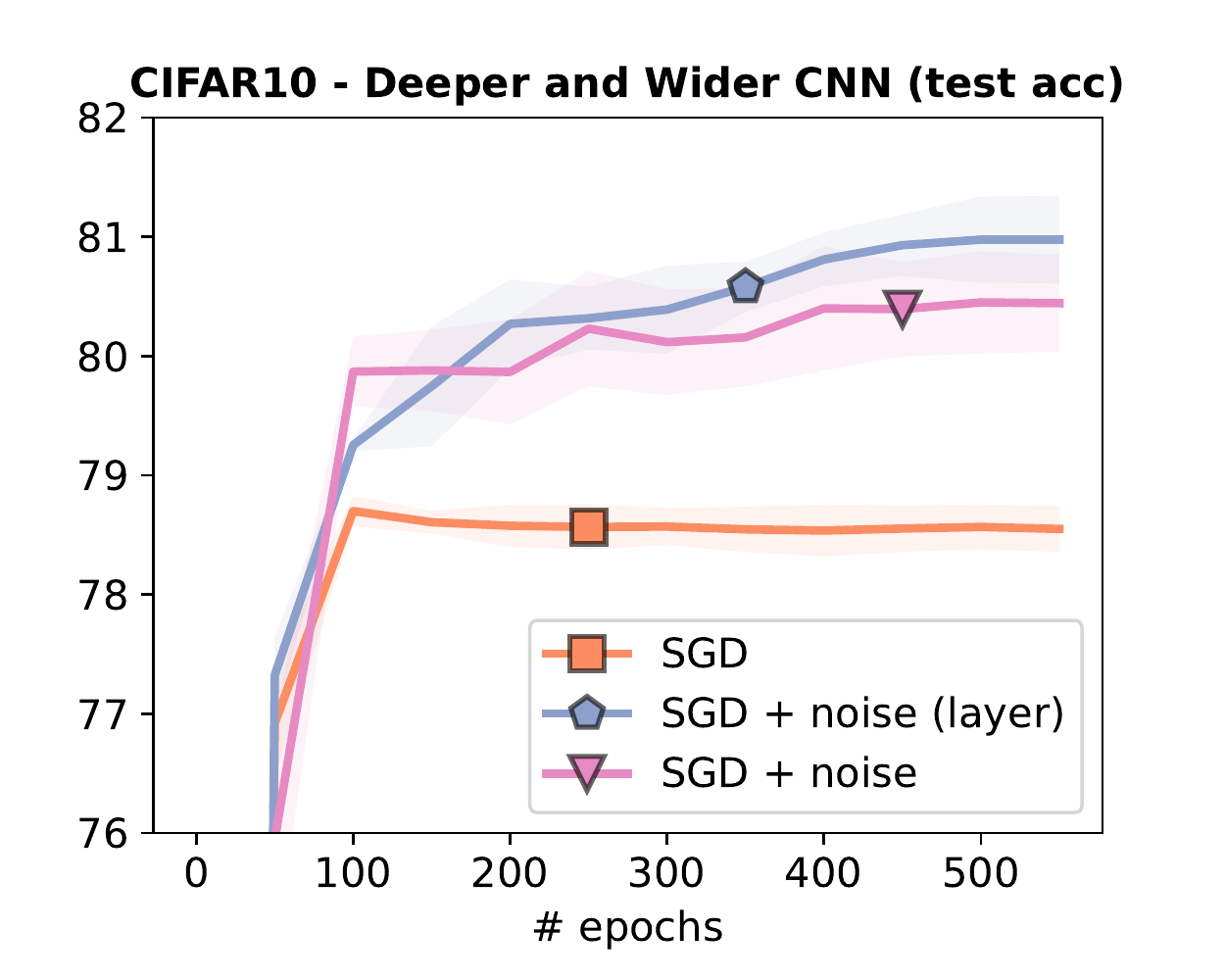}
    \includegraphics[height = 0.20\textwidth]{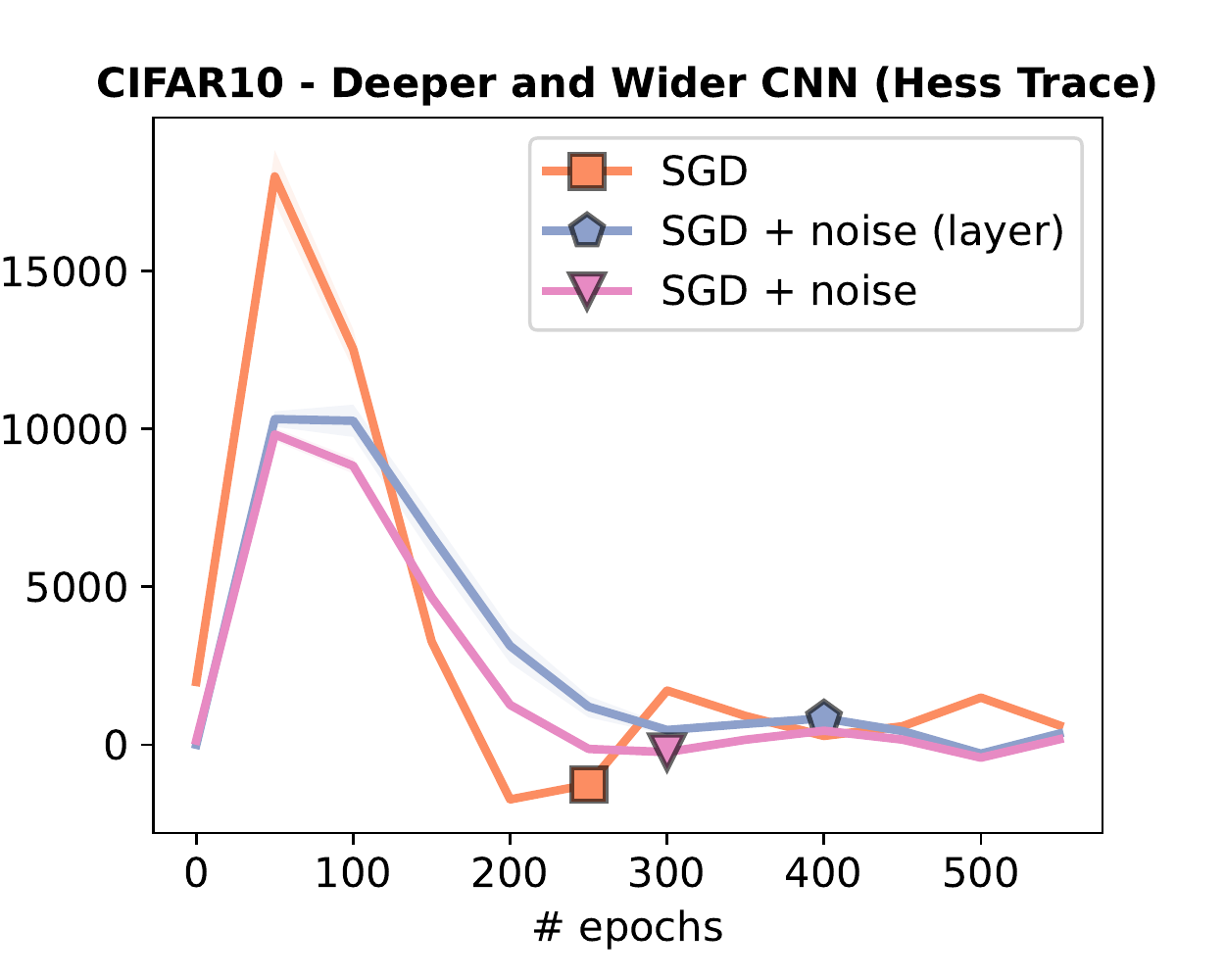}
    
    \vspace*{-.3cm}
    
    \caption{\small Behavior of noise injection in convolutional networks with different degrees of overparametrization. On the two \textbf{left} plots we show the performance using a toy PyTorch CNN+MLP~(5 channels) while on the two \textbf{right} plots we perform the same task on a fully convolutional network with 5 layers and 128 channels on 3 layers. All details are reported in the Appendix~\ref{ap:further_expts}. We run SGD with batch size $1024$ and learning rate $0.005$~($0.001$ for the PyTorch CNN+MLP) and cosine annealing~\citep{loshchilov2016sgdr}. We plot the performance for $\sigma =0.005$ of the two noise injection methods. Plots of the loss dynamics are provided in Appendix~\ref{ap:further_expts}.}
    \label{fig:CIFAR_summary}
    \vspace{-2mm}
\end{figure*}

\textbf{Small Convolutional Networks without normalization.} In Fig.~\ref{fig:CIFAR_summary} we go a bit beyond our theoretical setting and test the application to convolutional networks on CIFAR10~\citep{krizhevsky2009learning}, with full data and stochastic gradients~(batch-size $1024$). In this setting, we use a Toy CNN~(2 conv layers with 5 channels + MLP, details in the App), and compare against a wider fully-convolutional network with moderate width~(4 conv layers with 128 channels + one linear layer). The pattern we discussed above can be observed in this setting, albeit less pronounced due to the narrow nature of convolutional networks with reasonable number of channels. Note that, in the fully convolutional wide network, the trace of the Hessian in the noise injection setting is damped but oscillates around zero. This is quite different from the MLP setting.

\textbf{Deep Residual Networks.} To conclude, we test noise injection~(layer-wise or in all layers simultaneously) on a ResNet18~(around 11M parameters)~\citet{he2016deep} with batch normalization. We use for this the basiline SGD configuration in the popular repository \url{https://github.com/kuangliu/pytorch-cifar}. On this baseline, which reaches around 94.4\% test accuracy, we simply add noise~\footnote{Here by ``layers'' we mean each network parameter group. That is, noise is also injected in the batch-norm parameters.} injection at every step. In Figure~\ref{fig:CIFAR_big} we tested different noise injection standard deviations $\sigma$ and plotted mean and standard error of the mean~(3 runs) for the final test accuracy and Hessian trace after 150 epochs. We stopped noise injection at epoch 135 to allow the networks to converge and use cosine annealing, batch-size 128 and learning rate 0.01 in all methods. The results clearly show that injecting noise in all parameters is able to only regularize the objective and improve test accuracy for very small $\sigma$, but for $\sigma>0.1$ it hurts performance. Instead, injecting layer-wise noise provides a much more consistent regularization, and is able to improve performance by $+0.3\%$, which is a sizable margin given the strong SGD baseline. The poor performance of standard noise injection is predicted by the theory in Section~\ref{sec:neural-net}, which explains the sharp increase of the Hessian trace (for SGD + noise) observed in the second panel at $\sigma= 0.1$. 

\section{Conclusion}
\vspace{-2mm}
In this paper, we showed how explicit regularizers could be obtained from small random perturbations of parameters, both for simple machine learning models, and deep learning models with potentially a large number of neurons. This regularization is obtained by independent layer-wise perturbations, can be easily added to existing training codes, and lead to improved accuracies in our experiments.

Several avenues for future research naturally open: (a) obtain non-asymptotic optimization convergence results for the models in Section~\ref{sec:finitedim}, (b) a detailed theoretical study of the regularization properties of our new regularizer for deep networks, and (c) performance ablations on other architectures such as transformers~\citep{vaswani2017attention}.

\bibliographystyle{plainnat}
\bibliography{opt-ml}

\appendix
%

%

\onecolumn
\begin{center}
{ \Large{Explicit Regularization in Overparametrized Models via Noise Injection: \\
Supplementary Materials}}\\
\end{center}
\section{Proofs of equivalences} \label{ap:proof_theorems}
Note that in finite-dimensional models, all norms are equivalent, so we use any of them, unless otherwise stated.

    

\subsection{Proof of Theorem~\ref{theo:cross}} 



\begin{proof}
    We first observe from Eq.~\eqref{eq:def_Rsigma} that
    \begin{equation}
        |R_\sigma(w) -  R(w)|
        =
        \frac{\sigma^2}{2} {\rm D}  L(\Phi(w)) \underbrace{{\rm D}^2 \Phi(w) [  \idm ]}_{=0}  
        + \frac{\sigma^2}{2} {\rm D^2} L(\Phi(w)) \big[ {\rm D}\Phi(w) {\rm D}\Phi(w)^\top \big] + O(\sigma^3).
    \end{equation}
    By using the inequality (due to bounded second derivative of $\Phi$)
    \begin{equation} \label{eq:bound_in_proof_of_Thm1}
        {\rm D}\Phi(w)
        \leq 
        {\rm D}\Phi(0) + C \|w\|
        \leq 
        C(1+\|w\|),
    \end{equation}
    we obtain the first desired bound
    \begin{equation}
        \label{eq:proof_thm_1_2nd_bound}
        |R_\sigma(w) -  R(w)|
        =
        \left \vert \frac{\sigma^2}{2} \underbrace{{\rm D^2} L(\Phi(w))}_{\text{uniformly bounded by Ass.}} \big[ {\rm D}\Phi(w) {\rm D}\Phi(w)^\top \big] \right \vert
        \leq 
        C(1 + \| w \|^2) \sigma^2.
    \end{equation}
    For the second desired bound, we observe from Eqs.~\eqref{eq:noise_taylor} and \eqref{eq:second_order} that
    \begin{equation}
        |R_\sigma(w) -  R_\sigma^{({\rm eff})}(w)|
        =
        \left | \frac{\sigma^2}{2} {\rm D}  L(\Phi(w)) \underbrace{{\rm D}^2 \Phi(w) [  \idm ]}_{=0} + O(\sigma^3) \right |,
    \end{equation}
    which concludes the proof. (Note that in this proof the precise value of the constant $C>0$ is changing form line to line. It is not necessarily the same as the $C$ in the statement of the theorem.)
\end{proof}

\subsection{Proof of Theorem~\ref{theo:over}}

\begin{proof}
    As proved in Eq.~\eqref{eq:bound_in_proof_of_Thm1} under the same assumptions, we again have
    \begin{equation}
        \left \vert \frac{\sigma^2}{2} {\rm D^2} L(\Phi(w)) \big[ {\rm D}\Phi(w) {\rm D}\Phi(w)^\top \big] \right \vert
        \leq 
        C(1 + \| w \|_2^2)\sigma^2.
    \end{equation}
    (Here, and in the rest of the proof, $C>0$ is a constant independent of $\sigma$ which may change from line to line.)
    Similarly, by both the boundedness of second derivatives of $L$ and $\Phi$, we have that
    \begin{equation}
        \left \vert \frac{\sigma^2}{2} {\rm D}  L(\Phi(w)) {\rm D}^2 \Phi(w) [  \idm ] \right \vert \leq  C(\sigma^2(1 + \|\Phi(w) \|_2)).
    \end{equation}
    Thus, by putting both bounds together, there exists a constant  $C>0$ such that for all $w$, 
    \begin{align}\begin{split}
    |R_\sigma(w) - R(w)| 
    &\leq 
    \frac{\sigma^2}{2} {\rm D^2} L(\Phi(w)) \big[ {\rm D}\Phi(w) {\rm D}\Phi(w)^\top \big]
    +
    \frac{\sigma^2}{2} {\rm D}  L(\Phi(w)) {\rm D}^2 \Phi(w) [  \idm ] \\
    &\leq  C\sigma^2 ( 1 + \|\Phi(w)\|_2 + \|w \|_2^2) .
    \end{split}
    \end{align}
    Hence, if $w^{\sigma}_\ast$ is a minimizer of $R_\sigma$,  we have the chain of inequalities
    \[ R(w^{\sigma}_\ast) -C\sigma^2 (
    1 + \|\Phi(w^{\sigma}_\ast)\|_2 + \|w^{\sigma}_\ast \|_2^2) 
    ) \leq  R_\sigma(w^{\sigma}_\ast)
    \leq  R_\sigma (w_\ast) \hspace*{3cm}\]
    \[ \hspace*{4cm}  R_\sigma (w_\ast) 
    \leq  R (w_\ast) +C\sigma^2 ( 1 + \|\Phi(w_\ast)\|_2 + \|w_\ast \|_2^2).\]
    This leads to
    \[ \!\!
    \frac{\mu}{2} \|\Phi(w^{\sigma}_\ast) - \varphi_\ast\|_2^2 \leq  R(w^{\sigma}_\ast) -  R (w_\ast)
    \leq  2 C \sigma^2 ( 2 +\|\Phi(w_\ast)\|_2 
    + \|\Phi(w^{\sigma}_\ast)\|_2 
    +\|w^{\sigma}_\ast \|_2^2
    +\|w_\ast \|_2^2) \] by $\mu$-strong-convexity of $L$.
    Thus, since $w_\ast$ and $w_\ast^\sigma$ are in a predefined compact set $\Omega$ for $\sigma$ small enough and since $\Phi(w)$ is bounded on $\Omega$ due to boundedness of its second derivative, we arrive at:
     $\|\Phi(w^{\sigma}_\ast) - \varphi_\ast\|_2^2 =O(\sigma^2)$. This shows the first statement.
     
     We also consider a minimizer 
     $w^{\sigma, (\rm eff)}_\ast$ of $R_\sigma^{({\rm eff})}$, which satisfies the same bound as above (exact same reasoning), that  is, 
     \begin{equation}\label{eq:aux_eq_123}
     \|\Phi(w^{\sigma, (\rm eff)}_\ast) - \varphi_\ast\|^2 \leq  C'' \sigma^2.
     \end{equation}
     
     Since  
     \begin{equation*}
         \| {\rm D}  L(\Phi(w^{\sigma, (\rm eff)}_\ast)) \| \leq  C
     \end{equation*}
     due to $w^{\sigma, (\rm eff)}_\ast \in \Omega$ (compact set) 
     and 
     \begin{equation*}
         \| {\rm D}^2 \Phi(w^{\sigma, (\rm eff)}_\ast) \|
         = 
         \| {\rm D}^2 \Phi(w^{\sigma, (\rm eff)}_\ast) - {\rm D}^2 \Phi(\varphi_{\ast}) \|
         \leq
         C\|\Phi(w^{\sigma, (\rm eff)}_\ast) - \varphi_\ast\| \leq  C'' \sigma^2
     \end{equation*}
     due to Eq.~\eqref{eq:aux_eq_123}, we get
     \[ \frac{\sigma^2}{2} {\rm D}  L(\Phi(w^{\sigma, (\rm eff)}_\ast)) {\rm D}^2 \Phi(w^{\sigma, (\rm eff)}_\ast) [  \idm ]  \leq  C \sigma^3.\]
     This implies moreover that 
     \[ | R_\sigma^{({\rm eff})}( w^{\sigma, (\rm eff)}_\ast) - R_\sigma( w^{\sigma, (\rm eff)}_\ast)| \leq  C \sigma^3.\] 
     Moreover,  since $w_\ast^\sigma$ minimizes $R_\sigma$, the differential of $R_\sigma$ at $w_\ast^\sigma$ is equal to zero, and using a similar expansion than Theorem~\ref{theo:cross} for the differential, we get $0 = {\rm D}  L(\Phi(w_\ast^\sigma)) +
     O(\sigma)
     $. This leads to 
      $| R_\sigma^{({\rm eff})}( w_\ast^\sigma) - R( w_\ast^\sigma)| \leq  C \sigma^3$  due to $ {\rm D}  L(\Phi(w_\ast^\sigma)) \leq  C \sigma$. We can thus use the chain of inequalities 
      \[ R( w^{\sigma, (\rm eff)}_\ast)) -C\sigma^3  \leq   R_\sigma^{({\rm eff})}( w^{\sigma, (\rm eff)}_\ast))
    \leq   R_\sigma^{({\rm eff})} (w_\ast^\sigma)
    \leq  R (w_\ast^\sigma) +C\sigma^3,\] and follow the same reasoning as above to obtain the desired bound 
    $\|
    \Phi(w^{\sigma, (\rm eff)}_\ast) -  \Phi(w^{\sigma}_\ast) \|_2^2 = O(\sigma^{3})$.
\end{proof}
 
\subsection{Extension of asymptotic theory from Section \ref{sec:finitedim} to piecewise regularity of $\Phi$} 

In this section, we show how the derivation of $R_\sigma(w)$, Eq.~\eqref{eq:noise_taylor}, and Theorem~\ref{theo:cross} can be extended to deal with a more general predictor $\Phi$.
More precisely, we extend it to the case where globally $\Phi$ is only assumed to be continuous and the additional regularity of $\Phi$ is only given \emph{piecewise}, i.e., where there exists a finite number of open convex sets $\{\Omega_i,i=1,\dots,M\}$ with $\bigcup_{i \in I} \overline{\Omega_i} = \rb^m$ such that $\Phi$ restricted to $\Omega_i$ is in $C^3$ for all $i$. This is an important extension because it takes care of locally non-differentiable activation functions, such as ReLU.

\paragraph{Extension of Eq.~\eqref{eq:noise_taylor}}

Let $w \in \Omega_i$ for a fixed $i=1,\dots,M$.
Under the above piecewise assumption, the Taylor expansion of $\Phi$ at $w$ can now be written:
\begin{align}
     \Phi(w + \sigma \varepsilon) = &{\mathbbm{1}}_{w+\sigma \varepsilon \not\in \Omega_i} \big[ \Phi(w + \sigma \varepsilon)
    \big] \\ 
    &+ \mathbbm{1}_{w+\sigma \varepsilon \in \Omega_i} \big[ \Phi(w) + \sigma {\rm D}\Phi(w) \varepsilon + \frac{\sigma^2}{2} {\rm D}^2 \Phi(w) [ \varepsilon \varepsilon^\top] + O(\sigma^3 \| \varepsilon\|^3) \big].
\end{align}

Thus, as in the main text, we have
\begin{align*}
    L(\Phi(w+\sigma \varepsilon))
    =
    &\mathbbm{1}_{w+\sigma \varepsilon \not \in \Omega_i} \bigg[ 
    L(\Phi(w+\sigma \varepsilon))
    \bigg]  \\
    &+ \mathbbm{1}_{w+\sigma \varepsilon \in \Omega_i} \bigg[ \textstyle L (
 \Phi(w))  + DL(\Phi(w)) \big( \sigma {\rm D}\Phi(w) \varepsilon + \frac{\sigma^2}{2} {\rm D}^2 \Phi(w) [ \varepsilon \varepsilon^\top] + O(\sigma^3 \| \varepsilon\|^3)\big) \\
  &\hspace{1.3cm}+ 
 \frac{1}{2}{\rm D^2} L(\Phi(w)) \big[
 \sigma {\rm D}\Phi(w) \varepsilon (\sigma {\rm D}\Phi(w) \varepsilon)^\top + O(\sigma^3 \| \varepsilon\|^3) \big]
  + O(\sigma^3 \| \varepsilon\|^3) \bigg].
\end{align*}
Now, taking the expectation we get
\begin{align*}
    R_\sigma(w) &= \mathbb{E}\left[L(\Phi(w+\sigma \varepsilon)) \right] \\
    &=
    \int_{\{ \varepsilon: w+\sigma\varepsilon \not\in \Omega_i \}}
    L(\Phi(w+\sigma \varepsilon)) \mathrm{d} N(\varepsilon;0,\idm) \  + \
    \int_{\{ \varepsilon: w+\sigma\varepsilon \in \Omega_i \}}
    L(\Phi(w+\sigma \varepsilon)) \mathrm{d} N(\varepsilon;0,\idm) \\
    &=: J_1(\sigma) + J_2(\sigma),
\end{align*}
with the integrands from the above cases of $L(\Phi(w+\sigma \varepsilon))$.
Now,
\begin{equation*}
    \lim_{\sigma \to 0} J_2(\sigma)
    =
    L (
 \Phi(w)) +   \frac{\sigma^2}{2} {\rm D}  L(\Phi(w)) {\rm D}^2 \Phi(w) [  \idm ]  + \frac{\sigma^2}{2} {\rm D^2} L(\Phi(w)) \big[ {\rm D}\Phi(w) {\rm D}\Phi(w)^\top \big] + O(\sigma^3)
\end{equation*}
analogously to the main text, and
\begin{equation*}
    \lim_{\sigma \to 0} J_1(\sigma)
    =
    0
\end{equation*}
because the probability mass of $\{ \varepsilon: w+\sigma\varepsilon \not\in \Omega_i \}$ under $N(\varepsilon;0,\idm)$ goes to zero exponentially fast, while $L(\Phi)$ can diverge at most polynomially.
Thus, we arrive at Eq.~\eqref{eq:noise_taylor} from the main text.

\paragraph{Extension of Theorems~\ref{theo:cross} and \ref{theo:over}}

Above, we saw how our Taylor expansions and averaging carry over to the case where $\Phi$ is only piece-wise in $C^3$.
Holding this in mind, we can now verify that the above proofs of Theorems~\ref{theo:cross} and \ref{theo:over} also carry through in this case:

Since both Theorems are formulated only for sufficiently small $\sigma > 0$, this can be easily verified.
For Theorem~\ref{theo:cross} this is trivial, one just has to emphasize that $\sigma > 0$ is small enough after each step.
For Theorem~\ref{theo:over}, one additionally has to make sure that the minimizers $w_{\ast}^{\sigma}$ and $w^{\sigma, (\rm eff)}_\ast$ of $R(w) = L(\Phi(w))$ and $R_\sigma^{({\rm eff})}$ are also in $\Omega_i$. 
But this will hold true for all sufficiently small $\sigma$ for the following reason:
First, for $w_{\ast}^{\sigma}$ this follows from the fact that $J_2(\sigma)$ goes to $0$ exponentially fast, as $\sigma \to 0$.
Hence, only values of $L(\Phi(w))$ for $w \in \Omega_i$ matter to determine the minimizer $w_{\ast}^{\sigma}$, which will thus lie in $\Omega_i$.
Second, under the assumptions of Theorem~\ref{theo:over}, the difference between $R_\sigma$ and $R_\sigma^{({\rm eff})}$ is in $O(\sigma^3)$.
Thus, for $\sigma$ small enough, $w^{\sigma, (\rm eff)}_\ast$ will lie in the same open convex set as $w_{\ast}^{\sigma}$, i.e., in $\Omega_i$.

\section{Logistic regression} \label{ap:logistic}
We consider a 2-homogeneous model such that  $\Phi(\lambda w) = \lambda^2 \Phi(w)$ for any $w$ and $\lambda>0$. We aim to minimize
\[
\frac{1}{n} \sum_{i=1}^n \log(1 + \exp(-y_i \Phi(w)_i) ),
\]
where $y \in \{-1,1\}^n$. We assume that the model is overparameterized, so that there exists $w$ such that $y_i \Phi(w_i)>0$ for all $i$.

Following~\citet{lyu2019gradient}, we  expect $w$ to diverge in some direction, that is, $w = \lambda \Delta$ for $\lambda \to +\infty$ and $\|\Delta\|_2 = 1$. We assume that this is the case, and we derive here an informal argument highlighting what the limit direction $\Delta$ should be.  

In the function
\[
 \textstyle   R_\sigma^{({\rm eff})} (w) =  R(w) +  \frac{\sigma^2}{2} {\rm D^2} L(\Phi(w)) \big[ 
  {\rm D}\Phi(w) {\rm D}\Phi(w)^\top
 \big]
 \]
 from Equation~\eqref {eq:second_order},
the term ${\rm D}  \Phi(w) {\rm D}(\Phi(w))^\top$ grows  in $\lambda$ as $\lambda^{2} {\rm D}  \Phi(\Delta) {\rm D}(\Phi(\Delta))^\top$, while ${\rm D}^2 L(\Phi(w))$ is diagonal and proportional to $\exp( - \lambda^2 |\Phi(\Delta)_i|)$, with the same scaling as the loss function. We thus get an asymptotic approximate cost function equal to, since $| \Phi(\Delta)_i | = y_i \Phi(\Delta)_i$ (because we have perfect predictions):
\[
\frac{1}{n} \sum_{i=1}^n e^{-y_i \lambda^2\Phi(\Delta)_i }
\Big( 1   
+ \frac{\sigma^2}{2} \lambda^{2}
  \|{\rm D}(\Phi(\Delta))_i \|_2^2
\Big) \sim
\frac{1}{n} \sum_{i=1}^n e^{-y_i  \Phi(w)_i }
\Big( 1   
+ \frac{\sigma^2}{2}  
  \|{\rm D}(\Phi(w))_i \|_2^2
\Big).
\]
By taking the gradient with respect to $w$, we get that
\[
\sum_{i=1}^n e^{-y_i \Phi(w)_i}
\big( \frac{\sigma^2}{2} \frac{\partial}{\partial w} 
\|{\rm D}(\Phi(w))_i \|_2^2 - y_i \Big( 1   
+ \frac{\sigma^2}{2}  
  \|{\rm D}(\Phi(w))_i \|_2^2
\Big) \frac{\partial}{\partial w} \Phi(w)_i\Big),
\]
which is asymptotically equivalent to
\[
\sum_{i=1}^n e^{-y_i \Phi(w)_i}
\big( \frac{\sigma^2}{2} \frac{\partial}{\partial w} 
\|{\rm D}(\Phi(w))_i \|_2^2 - y_i  \frac{\partial}{\partial w} \Phi(w)_i\Big) \propto - w.
\]
We conjecture that this is equivalent to the optimality conditions of the problem
\begin{align*}
 \max_{\| \Delta\|_2 \leq  1}
\min_{i \in \{1,\dots,n\}} \Big\{
y_i \Phi(w)_i - \frac{\sigma^2}{2} \|{\rm D}(\Phi(w))_i \|_2^2
\Big\},
\end{align*}
where for $\sigma = 0$, this is the result of ~\citet{lyu2019gradient}. We leave a formalization of such a result for future work.

\section{Direct derivations of formulas from Sections~\ref{sec:lasso} and~\ref{sec:linear_networks}} \label{ap:formula_derivation}

This section contains some explicit derivation of formulas that were asserted in the main text. 

\subsection{Lasso} \label{app:subsec:lasso_derivation}

Here, we present the derivation of Equation~\eqref{eq:def_Lasso}. First we note that ${\rm D^2} L(\Phi(w)) = \frac 1n \idm$, i.e.~the identity matrix, and that
\begin{equation}
    {\rm D}\Phi(w) = \begin{bmatrix} 2X \diag(w_1), 2X \diag(w_2) \end{bmatrix}.
\end{equation}
Insertion of these two identities into Eq.~\eqref{eq:second_order} yields
\begin{align} \label{eq:R_sig_lasso}
    R_\sigma^{({\rm eff})} (w) 
    &=  
    R(w)  +  \frac{\sigma^2}{2} {\rm D^2} L(\Phi(w)) \big[ 
    {\rm D}\Phi(w) {\rm D}\Phi(w)^\top
    \big]
    \\
    &=
    R(w) +  2\sigma^2 \underbrace{\frac 1n \idm \Bigg[ \begin{bmatrix} X \diag(w_1), X \diag(w_2) \end{bmatrix} \begin{bmatrix} X \diag(w_1) \\  X \diag(w_2) \end{bmatrix} \Bigg]}_{=:J}.
\end{align}
By our above notation ${\rm D^2} L(\varphi) [ M ] = \sum_{a,b=1}^n {\rm D^2} L(\varphi)_{ab} M_{ab} \in \rb$, we have
\begin{equation}
    J = \frac 1n \sum_{a=1}^d \begin{bmatrix} X \diag(w_1^2) X^\top /n \end{bmatrix}_{aa} + \sum_{a=1}^d \begin{bmatrix} X \diag(w_1^2) X^\top /n \end{bmatrix}_{aa} := J_1 + J_2.
\end{equation}
We keep computing:
\begin{align}
    J_1 &= \sum_{a=1}^n \sum_{i=1}^d \frac 1n X_{ai} {w_1}_i^2 X_{ia}^\top 
    = \sum_{i=1}^d \underbrace{\left( \sum_{i=1}^n \frac 1n X_{ia}^\top X_{ai}  \right)}_{[X^\top X]_{ii}} {w_1}_i^2
    = \diag(X^\top X/n)^\top [w_1 \circ w_1],
\end{align}
and analogously $J_2 = \diag(X^\top X/n)^\top [w_2 \circ w_2]$.
Thus, $J = \diag(X^\top X/n)^\top [w_1 \circ w_1] + \diag(X^\top X/n)^\top [w_2 \circ w_2]$ which we insert back into Eq.~\eqref{eq:R_sig_lasso}.
This yields the desired Eq.~\eqref{eq:def_Lasso}.

\subsection{Nuclear norm (linear networks)}

For the main text, only the derivation of Eq.~\eqref{eq:R_sigma_nuclear_norm} is missing.
To this end, we first observe that
\begin{equation}
    [W_2 W_1 X^\top]_{ij}
    =
    \sum_{k=1}^{d_0} \sum_{l=1}^{d-1} [W_2]_{il} [W_1]_{lk} X_{jk}.
\end{equation}
Moreover,
\begin{equation}
    {\rm D^2} L(\Phi(w)) = \frac 1{2n} \idm.
\end{equation}
Hence,
\begin{align}
    {\rm D^2} L(\Phi(w)) \big[ {\rm D}\Phi(w) {\rm D}\Phi(w)^\top \big]
    &=
    \frac{1}{2n} \sum_{a=1}^{nd_2} \big[ {\rm D}\Phi(w) {\rm D}\Phi(w)^\top \big]_{aa} \\
    &= 
    \frac{1}{2n} \sum_{i=1}^{d_2} \sum_{j=1}^n 
    \left[ \left \Vert \nabla_w \sum_{k=1}^{d_0} \sum_{l=1}^{d_1} [W_2]_{il} [W_1]_{lk} X_{jk} \right \Vert_2^2 \right] \\
    &=
    \frac{1}{2n} \sum_{i=1}^{d_2} \sum_{j=1}^n  \left[  \sum_{k=1}^{d_0} \sum_{l=1}^{d_1} \left([W_2]_{il} X_{kj}^\top \right)^2    +  \sum_{l=1}^{d_1} \left ( \sum_{k=1}^{d_0} [W_1]_{lk} X^\top_{kj}   \right)^2 \right] \\
    &=
    \frac{1}{2n}
    (J_1 + J_2),
\end{align}
with
\begin{align}
    J_1 :=&
    \sum_{i=1}^{d_2} \sum_{j=1}^n \sum_{k=1}^{d_0} \sum_{l=1}^{d_1} [W_2]^2_{il} X_{jk}^2  \\
    =&
    \left [ \sum_{i=1}^{d_2} \sum_{l=1}^{d_1} [W_2]^2_{il} \right]^2 
    \left [ \sum_{j=1}^n \sum_{k=1}^{d_0} X_{jk}^2 \right ]^2 \\
    =&
    \Vert W_2 \Vert_F^2 \Vert X \Vert_F^2
\end{align}
and
\begin{align}
    J_2 :=
    \sum_{i=1}^{d_2} \sum_{j=1}^n \sum_{l=1}^{d_1} [ W_2 X^\top ]^2_{lj} =
    d_2 \Vert W_1 X^\top \Vert_F^2.
\end{align}
Putting all equations to together concludes the derivation of Eq.~\eqref{eq:R_sigma_nuclear_norm}.

\subsection{Group Lasso}

\paragraph{Extension to ``group Lasso''.} It is traditional to recover the group Lasso as a special of nuclear norm minimization \citep[see, e.g.,][]{bach2008consistency}.
We thus consider $w = (v_1,w_1,\dots, v_d,w_d) \in \rb^{d(k+1)}$,  with $v_j \in \rb$ and $w_j \in \rb^k$, and $\Phi(w) = v_1 X_1 w_1 + \cdots + v_d X_d w_d$, for $X_1,\dots,X_d \in \rb^{ n \times k}$, and  $L(\varphi) = \frac{1}{2n} \| y - \varphi \|_2^2$. This corresponds exactly to a linear network defined above, with $X = (X_1,\dots,X_n)$, $W_1$ defined by blocks and block-diagonal with blocks $w_j^\top$, and $W_2 = v^\top$.

Regardless whether the model is overparametrized or not, we can apply Theorem~\ref{theo:cross}, and we get: 
\begin{align} \label{eq:R_sigma_group_lasso}
    \begin{split}
    R_\sigma^{({\rm eff})} (w)   = &\frac{1}{2n} \| y -  v_1 X_1 w_1 - \cdots - v_d X_d w_d\|_2^2 \\ &+ 
    \frac{\sigma^2}{2n} \Big[ \| X_1 w_1\|_2^2 + v_1^2 \| X_1\|_F^2 + \cdots +  \| X_d w_d\|_2^2 + v_d^2 \| X_d\|_F^2 \Big].
    \end{split}
\end{align}
See the Appendix for a detailed derivation of Eq.~\eqref{eq:R_sigma_group_lasso}.
Optimizing over the ``invariance of scale'' (we can multiply $v_j$ by $\alpha_j$ and divide $w_j$ by the same $\alpha_j$), we get the equivalent problem of minimizing
\begin{equation}
    \frac{1}{2n} \| y -   X_1 \beta_1 - \cdots -   X_d \beta_d\|_2^2 \ + 
    \frac{\sigma^2}{n}  \Big[  \| X_1\|_F \cdot \| X_1 \beta_1\|_2  + \cdots +   \| X_d\|_F \cdot \| X_d \beta_d\|_2 \Big],
\end{equation}
where $\beta_j = |v_j| w_j$, 
which is a form of group Lasso. In particular if all $\beta_j$'s have dimension one, we recover the Lasso with a different formulation from Section~\ref{sec:lasso} (and no need for overparametrization).

\paragraph{Derivation of Equation~\eqref{eq:R_sigma_group_lasso}}
\label{app:derivation_group_lasso}

First we note that ${\rm D^2} L(\Phi(w)) = \frac 1n \idm$, i.e.~the identity matrix, and that
\begin{equation}
    \Phi(w) = \sum_{i=1}^d \Phi_i(w), 
    \quad \text{where }
    \Phi_i(v_i,w_i) := v_i X_i w_i .
\end{equation}
Under this notation, we get the Jacobian
\begin{equation}
    {\rm D}\Phi(w) = 
    \begin{bmatrix} {\rm D}\Phi_1(w) & & \\ & \ddots & \\  & & {\rm D}\Phi_d(w) \\       \end{bmatrix},
\end{equation}
with
\begin{equation}
    {\rm D}\Phi_i(v_i,w_i) =
    \begin{bmatrix} X_i w_i, v_i X_i \end{bmatrix} \in \R^{n \times (k+1)}.
\end{equation}
Now, 
\begin{equation}
    {\rm D}\Phi_i(v_i,w_i) {\rm D}\Phi_i(v_i,w_i)^\top
    =
    \begin{bmatrix} X_i w_i, v_i X_i \end{bmatrix} \begin{bmatrix} w_i^\top X_i^\top \\ v_i X_i^\top \end{bmatrix}
    =
    \begin{bmatrix} X_i w_i w_i^\top X_i^\top , v_i^2 X_i X_i^\top  \end{bmatrix},
\end{equation}
which implies that
\begin{align}
    {\rm D}\Phi(w) \cdot {\rm D}\Phi(w)^\top
    &=
    \begin{bmatrix}
        {\rm D}\Phi_1(w) {\rm D}\Phi_1(w)^\top & & \\
        & \ddots & \\
        & & {\rm D}\Phi_d(w) {\rm D}\Phi_d(w)^\top
    \end{bmatrix} \\
    &=
    \begin{bmatrix}
        \begin{bmatrix} X_1 w_1 w_1^\top X_1^\top , v_1^2 X_1 X_1^\top  \end{bmatrix} & & \\
        & \ddots & \\
        & & \begin{bmatrix} X_d w_d w_d^\top X_d^\top , v_d^2 X_d X_d^\top  \end{bmatrix}
    \end{bmatrix}
    .
\end{align}
By our above notation ${\rm D^2} L(\varphi) [ M ] = \sum_{a,b=1}^n {\rm D^2} L(\varphi)_{ab} M_{ab} \in \rb$, we now get the desired formula:
\begin{align}
    R_\sigma^{({\rm eff})} (w) 
    &=  
    R(w)  +  \frac{\sigma^2}{2} {\rm D^2} L(\Phi(w)) \big[ 
    {\rm D}\Phi(w) {\rm D}\Phi(w)^\top
    \big]
    \\
    &=
    R(w) +  \frac{\sigma^2}{2n} I [{\rm D}\Phi(w) \cdot {\rm D}\Phi(w)^\top] \\
    &=
    R(w) + \frac{\sigma^2}{2n} [ \Vert X_1 w_1 \Vert_2^2 + v_1^2 \Vert X_1 \Vert_F^2 + \dots + \Vert X_d w_d \Vert_2^2 + v_d^2 \Vert X_d \Vert_F^2 ].
\end{align}

\section{Appendix for Section~\ref{sec:neural-net} (one-hidden layer)} \label{ap:sec_3_neural_net}

\subsection{Explosion of full perturbations}
We have discussed the exploding variance phenomena in one hidden layer linear networks in Section~\ref{sec:explosion}. We here discuss the variance explosion in one hidden layer neural networsk with ReLU activations . 
Similarly, we here consider the overparametrized limit $d_1 \to +\infty$ with initialization   which corresponds to having weights of order $(W_1)_{ij} \sim \frac{1}{\sqrt{d_1 d_0}}$ for all $i,j$, that is $\|W_1\|_F^2$ not exploding with $d_1$, and $(W_2)_{ij} \sim \frac{1}{\sqrt{d_2 d_1}}$, that is $\|W_2\|_F^2$ not exploding with $d_1$. This initialization corresponds to the one in ~\cite{glorot2010understanding} in the case where the network was constant width. We have 
\begin{align*}
    \Phi(w) = W_2 (W_1X^\top)_{+},
\end{align*}
where the positive part is taken element-wise.

We have, with Gaussian perturbations $E_1$ and $E_2$, an explicit exact expansion for one hidden layer ReLU networks for small $\sigma$ (so that $\Phi$ is locally quadratic):
\BEAS
& & \Phi(w+ \sigma \varepsilon)  \\
& = &  \Phi(  W_{2}+\sigma E_2, W_1+\sigma E_1) =   (W_{2}+\sigma E_2)[(W_1+\sigma E_1) X^\top]_{+} \\
&=&(W_{2}+\sigma E_2)(W_1X^\top)_{+} + \sigma (W_{2}+\sigma E_2) \big[( W_1 X^\top)_+^0 \circ E_1 X^\top \big]  \\
&=& W_2 (W_1X^\top)_{+} + \sigma E_2 (W_1X^\top)_{+} + \sigma W_2\big[( W_1 X^\top)_+^0 \circ E_1 X^\top \big]  + \sigma^2 E_2\big[( W_1 X^\top)_+^0 \circ E_1 X^\top \big]  \\
  & = &  \Phi(w) + \sigma \big(  E_2 (W_1X^\top)_{+}  + W_2\big[( W_1 X^\top)_+^0 \circ E_1 X^\top \big] \big)  + \sigma^2  E_2\big[( W_1 X^\top)_+^0 \circ E_1 X^\top \big]. 
\EEAS
Taking expectations and using that $E_1,E_2$ have zero mean and are independent, and such that, $\E [ E_i M E_i^\top ] = \tr(M) \idm $ for $i=1,2$, and $M$ any symmetric matrix of compatible size, we can get $\E \big[ \Phi(w+ \sigma \varepsilon)  \big]
 = \Phi(w)$, and:
 
\BEAS
& & \E \big[ \|  \Phi(w+ \sigma \varepsilon)\|_F^2 \big] \\
   & = &  \| \Phi(w)\|_F^2 +   {\sigma^2}  \| (W_1 X^\top)_+ \|_F^2+  {\sigma^2}  \sum_{j=1}^{d_1} \sum_{i=1}^n \| (W_2)_{\cdot j}\|_2^2 \times 
| (( W_1 X^\top)_+^0)_{ji} |^2  \times \|X_{i\cdot}\|_2^2    \\
& & \hspace{3cm} + \sigma^4  \E \left \langle E_2\big[( W_1 X^\top)_+^0 \circ E_1 X^\top \big],E_2\big[( W_1 X^\top)_+^0 \circ E_1 X^\top \big] \right \rangle .
\EEAS

Consider a case when $(W_1X^\top)_{+}^0  =1$ at all of its entries. In that case, 
We can now compute $R_\sigma$ as:
\BEAS
\!R_\sigma(W_1,W_2) &  = &   R(W_1,W_2) 
+ \frac{\sigma^2}{2n} \big[ \| (W_1 X^\top)_+ \|_F^2  \\
& & + \frac{\sigma^2}{2n} \sum_{j=1}^{d_1} \sum_{i=1}^n \| (W_2)_{\cdot j}\|_2^2 \times 
| (( W_1 X^\top)_+^0)_{ji} |^2 
\times \|X_{i\cdot}\|_2^2     + \frac{\sigma^4}{2n}  
 d_1 d_2 \| X\|_F^2
.
\EEAS
The extra term $\frac{\sigma^4}{2n} d_2 d_1 \| X\|_F^2$  is of superior order in $\sigma$, but problematic  when $d_1 \to +\infty$ using same argument as that for linear net with one hidden layer. 



\subsection{Layer-wise perturbation}
\label{sec:laypert}
Now, we show that layer wise perturbation helps in variance control. 
Consider the following two cases : 
\begin{align*}
    \Phi(W_2,W_1+\sqrt{2}\sigma E_1) &= W_2 [(W_1+\sqrt{2}\sigma E_1) X^\top]_{+} = W_2 (W_1X^\top)_{+} + \sqrt{2}\sigma W_2\big[( W_1 X^\top)_+^0 \circ E_1 X^\top \big] \\
    &=\Phi(W_2,W_1)+ \sqrt{2}\sigma W_2\big[( W_1 X^\top)_+^0 \circ E_1 X^\top \big].
\end{align*}
Similarly, 
\begin{align*}
    \Phi(W_2+\sqrt{2}\sigma E_2,W_1) & = (W_2+\sqrt{2}\sigma E_2) (W_1X^\top)_{+} = W_2 (W_1X^\top)_{+} + \sqrt{2}\sigma E_2 (W_1X^\top)_{+} \\
    &=\Phi(W_2,W_1)  + \sqrt{2}\sigma E_2 (W_1X^\top)_{+}.
\end{align*}
Hence, 
\begin{align*}
    \E \|\Phi(W_2,W_1+\sqrt{2}\sigma E_1)  \|_F^2  = \|\Phi(W_2,W_1)  \|_F^2 + 2 \sigma^2\sum_{j=1}^{d_1} \sum_{i=1}^n \| (W_2)_{\cdot j}\|_2^2 \times 
| (( W_1 X^\top)_+^0)_{ji} |^2  \times \|X_{i\cdot}\|_2^2 \big].
\end{align*}
Similarly,
\begin{align*}
     \E \|\Phi(W_2+\sqrt{2}\sigma E_2,W_1)  \|_F^2  =\|\Phi(W_2,W_1)  \|_F^2 + 2 \sigma^2 \| (W_1 X^\top)_+ \|_F^2.
\end{align*}
If we choose uniformly at random the layer to be perturbed, then one can then see that\footnote{Note that in the main paper, there is a typo, and the term $\|X_{i\cdot}\|_2^2 $ was missing.}
\BEAS
\!R_\sigma(W_1,W_2)  & = &   R(W_1,W_2) 
+ \frac{\sigma^2}{2n} \big[ \| (W_1 X^\top)_+ \|_F^2  \\
& & + \frac{\sigma^2}{2n} \sum_{j=1}^{d_1} \sum_{i=1}^n \| (W_2)_{\cdot j}\|_2^2 \times 
| (( W_1 X^\top)_+^0)_{ji} |^2 \times \|X_{i\cdot}\|_2^2  \big].
\EEAS
Due to same argument as that for one hidden layer linear net, $R_{\sigma}(W_1,W_2)$ is well behaved.


\subsection{Function space}

Here we change slightly the notations to be closer to notations from~\citet{chizat2020implicit}.

Given some data $x_1,\dots,x_n$, we consider the function $\psi(a,b) = a ( b^\top x_i)_+$, with $(a,b) \in \rb^{d+1}$, and $\Phi(a_1,b_1,\dots,a_m,b_m) = \frac{1}{m} \sum_{i=1}^n \psi(a_i,b_i)$, with all weights with unit scale. This corresponds exactly to the one-hidden layer neural network with Glorot initialization, where $m = d_1$.

We consider $L(\varphi) = \frac{1}{2n} \| y - \varphi\|_2^2$, and the equivalent cost function that we obtained is exactly
 $$
  \frac{1}{2n} \sum_{i=1}^n 
 \Big(
 y_i - \frac{1}{m} \sum_{j=1}^m a_j (b_j^\top x_i)_+
 \Big)^2 + \frac{\sigma^2}{2nm} \sum_{i=1}^n \sum_{j=1}^m \Big\{
 (b_j^\top x_i)_+^2 + a_j^2 \|x_i\|_2^2 ( b^\top x_i)_+^0
 \Big\}.
 $$
  We can optimize over the scale $\lambda_j > 0 $ so that $a_j \rightarrow a_j \lambda$ and $b_j \rightarrow b_j  / \lambda$, leading to
  $$
\frac{1}{2n} \sum_{i=1}^n 
 \Big(
 y_i - \frac{1}{m} \sum_{j=1}^m a_j (b_j^\top x_i)_+
 \Big)^2 + \frac{\sigma^2}{m}  \sum_{j=1}^m  \sqrt{\frac{1}{n} \sum_{i=1}^n
|a_j|^2  \|x_i\|_2^2 (b_j^\top x_i)_+^0} \sqrt{\frac{1}{n}\sum_{i=1}^n
   (b_j^\top x_i)_+^2} ,
 $$
that is,
 $$
  \frac{1}{2} \E \Big( y - \int a (b^\top x)_+ d\mu(a,b) \Big)^2
 + \sigma^2 \int |a| \sqrt{ \E  \|x\|_2^2 (b^\top x)_+^0} \sqrt{ \E(b^\top x)_+^2}  d\mu(a,b),
 $$
 where $\E$ is the empirical expectation over the data and $d\mu(a,b) = \frac{1}{m} \sum_{j=1}^m \delta_{(a_j,b_j)}$.
  Denoting $d\nu(b) = a d\mu(a,b)$, we get the following const function with respect to $\nu$:
  $$
 F_\sigma(\nu) =\frac{1}{2} \E \Big( y - \int   (b^\top x)_+ d\nu(b) \Big)^2
 + \sigma^2 \int   \sqrt{ \E \|x\|_2^2 (b^\top x)_+^0} \sqrt{ \E(b^\top x)_+^2} |d\nu(b)|.
 $$
  We thus get a specific $\ell_1$-penalty, where the difference with the variation norm of~\citet{kurkova,bach2017breaking} is the presence of the data-dependent  multiplier 
  $ \sqrt{ \E  \|x\|_2^2 (b^\top x)_+^0} \sqrt{ \E(b^\top x)_+^2}$, rather than simply having a constant. With sufficiently many data points, these expectations are bounded from below and above, so the regularization properties are the same, with the same adaptivity to linear substructures highlighted by~\citet{chizat2020implicit}.

\section{Appendix for Section~\ref{sec:neural-net-deep} (deep networks)} \label{ap:sec_4_neural_net_deep}

\subsection{Explosion of full perturbations}
\subsubsection{Linear Network}
Consider a linear neural network with $w = (W_1,,\dots, W_M )$ with $W_i \in \mathbb{R}^{d_i \times d_{i-1}}$, and  $\Phi(w) = W_M W_{M-1} \cdots W_1 X^{\top} \in \rb^{d_M \times n}$ with the input data $X\in \mathbb{R}^{n\times d_0}$, with $L(\varphi) = \frac{1}{2n}\|Y^\top - \varphi \|_F^2$ for output data $Y \in \rb^{n \times d_M}$ and $\varphi \in \rb^{ d_M \times n}$.  Let us denote the noise with $\varepsilon =(E_1,E_2,\dots,E_M)$. Let us now compute,
\begin{align*}
    \Phi(w+\sigma \varepsilon) &= \Phi(W_M+\sigma E_M, \dots, W_1+\sigma E_1) =(W_M+\sigma E_M)\dots(W_1+\sigma E_1)X^\top \\
    &=W_M\dots W_1X^\top + \sigma^M E_M\dots E_1X^\top + \sum_{i=1}^{M-1} \sigma^{i} C_i \\
    &= \Phi(W_M,\dots,W_1) + \sigma^M E_M\dots E_1X^\top + \sum_{i=1}^{M-1} \sigma^{i} C_i,
\end{align*}
where $C_i$ depends on $W_M,\dots,W_1$ and $E_M,\dots,E_1$. It is easy to check that $\E[\Phi(w+\sigma \varepsilon)] = \Phi(w)$. To show that $R_\sigma$ explodes we will show that $ \|\Phi(w+\sigma \varepsilon)\|_F^2$ explodes. Because of independence of $E_1,\dots E_M$, we have
\begin{align*}
    \E\| \Phi(w+\sigma \varepsilon)\|_F^2 = \|\Phi(W_M,\dots,W_1)\|_F^2 + \sigma^{2M} \E \| E_M\dots E_1X^\top\|_F^2 + \sum_{i=1}^{M-1} \sigma^{2i} \E \|C_i\|_F^2.
\end{align*}
Consider the term
\begin{align*}
    \sigma^{2M} \E \| E_M\dots E_1X^\top\|_F^2 = \sigma^{2M} d_1 d_2\cdots d_M \|X\|_F^2.
\end{align*}
In the above we used the fact that $\E[E_i M E_i^\top] = \tr(M) I$.
As $d_i \rightarrow \infty$ for $i \in \{1,\dots,M-1\}$, $\sigma^{2M} \E \| E_M\dots E_1X^\top\|_F^2$ diverges and hence $R_{\sigma}$ explodes as well. Using the same argument, we can show that all coefficients $\sigma^{2i}$ for all $i >1$ explode. 

\subsubsection{ReLU Network}
Consider a deep neural network with $w = (W_1,,\dots, W_M )$ with $W_i \in \mathbb{R}^{d_i \times d_{i-1}}$, and  $\Phi(w) = W_M W_{M-1} \cdots W_1 X^{\top} \in \rb^{d_M \times n}$ with the input data $X\in \mathbb{R}^{n\times d_0}$, with $L(\varphi) = \frac{1}{2n}\|Y^\top - \varphi \|_F^2$ for output data $Y \in \rb^{n \times d_M}$ and $\varphi \in \rb^{ d_M \times n}$. We use ReLU activation here. Let us denote the noise with $\varepsilon =(E_1,E_2,\dots,E_M)$. We have,
\begin{align*}
    \Phi(W_M,\dots, W_1) =  W_M(W_{M-1}\dots (W_1X^\top)_{+}\dots )_{+}
\end{align*}
Now, let us compute
 
\begin{align*}
    &\Phi(W_M+\sigma E_M,\dots, W_1+\sigma E_1) = (W_M+\sigma E_M)((W_{M-1}+\sigma E_{M-1})\dots ((W_1+\sigma E_1)X^\top)_{+}\dots )_{+} \\
    &=W_M(W_{M-1}\dots (W_1X^\top)_{+}\dots )_{+}+ + \sum_{i=1}^{M-1} \sigma^{i} C_i \\
    &\hspace{3cm}+ \sigma^{M} E_M((W_{M-1}\dots (W_1X^\top)_{+}\dots )_{+}^0 \circ E_{M-1}(\dots ( W_1 X^\top)_+^0 \circ E_1 X^\top)) ,
\end{align*}
where $C_i$ depends on $W_M,\dots,W_1$ and $E_M,\dots,E_1$. Using similar argument as for linear network,
\begin{align*}
    \E\|\Phi(w+\sigma \varepsilon)\|_F^2  = \|\Phi(w)\|_F^2  &+ \sigma^{2M} \E \|E_M((W_{M-1}\dots (W_1X^\top)_{+}\dots )_{+}^0 \circ E_{M-1}(\dots ( W_1 X^\top)_+^0 \circ E_1 X^\top)) \|_F^2 \\
    &\hspace{4cm}+ \sum_{i=1}^{M-1} \sigma^{2i} \E \|C_i\|_F^2.
\end{align*}

Let us again consider a special case and assume that $w = (W_M,\dots,W_1)$ are such that $W_1X^\top >0$ at all entries and $W_2,\dots,W_M$ has all positive entries. In such scenario, for small enough $\sigma$
\begin{align*}
    &\sigma^{2M} \E \|E_M((W_{M-1}\dots (W_1X^\top)_{+}\dots )_{+}^0 \circ E_{M-1}(\dots ( W_1 X^\top)_+^0 \circ E_1 X^\top)) \|_F^2 \\ &= \sigma^{2M} \E\| E_M\dots E_1X^\top \|_F^2 =\sigma^{2M} d_1\cdots d_M \|X\|_F^2.
\end{align*}
In the above we used the fact that $\E[E_i M E_i^\top] = \tr(M) I$.
As $d_i \rightarrow \infty$ for $i \in \{1,\dots,M-1\}$, $\sigma^{2M} \E \| E_M\dots E_1X^\top\|_F^2$ diverges and hence $R_{\sigma}$ explodes as well. We can use similar constructions to show that all coefficients $\sigma^{2i}$ for $i >1$ explode.


\subsection{Layer-wise perturbation}
In this section, we will show that how layer wise perturbations leads to the desired result. We then proceed to show that the term is not exploding.

The perturbation argument follows exactly the same line as Section~\ref{sec:laypert}, using that neural networks with ReLU activations or without activations are positively homogeneous with respect to each layer.


We thus simply need to show that the extra term is not exploding. For linear networks, the regularization term is
$$\frac{\sigma^2}{2n}
\sum_{j=1}^M
 \big\| W_M \cdots W_{j+1} \big\|_F^2  \big\| W_{j-1} \cdots W_1 X^\top\big\|_F^2,
$$
and with scaling described in the paper, we get that each element of
 $W_M \cdots W_{j+1} $ is the product of
 $d_{M-1} \cdots d_{j+1} $ terms, each of scale\footnote{We recall that this initialization, i.e. each weight between layer $i$ and $i-1$ with variance $1/\sqrt{d_id_{i-1}}$, which stabilizes the expected Frobenius norm of the weight matrices. } $\frac{1}{\sqrt{d_{M} d_{M-1}}} \cdots \frac{1}{\sqrt{d_{j+1} d_j}}$, thus we an overall scale of 
 $$
 d_{M-1} \cdots d_{j+1}
 \times \frac{1}{\sqrt{d_{M} d_{M-1}}} \cdots \frac{1}{\sqrt{d_{j+1} d_j}}
 = \frac{1}{\sqrt{d_M d_j}}.
 $$
 Then, the Frobenius norm of $W_M \cdots W_{j+1} $ is of order 
 $d_M d_j \times \big(  \frac{1}{\sqrt{d_M d_j}} \big)^2 = 1$.
The reasoning is the same for $\big\| W_{j-1} \cdots W_1 X^\top\big\|_F^2$ and leads to no explosion.

With ReLU activation functions, since the derivative of the ReLU activation is either 0 or 1, this does not change the scalings.

\section{Further experimental details}
\label{ap:further_expts}
\label{ap:further_exp}
\vspace{-3mm}
We provide a thorough overview of the experimental results presented in this work. Some of the plots of the main paper are also presented again here for ease of comparison.
    \vspace{-3mm}
\paragraph{Datasets and networks.} We use the Fashion MNIST~\citep{xiao2017fashion} and CIFAR10~\citep{krizhevsky2009learning} datasets and train on different neural network models~(6 experiments in total). 
    \vspace{-3mm}
\paragraph{Setup.} We repeat all experiments 3 times~(different network initializations and injected noise) and show mean and 1 standard deviation. All experiments are conducted in PyTorch~\citep{paszke2017automatic} on up to 8
Tesla V100 GPUs with 32 GB memory. 
    \vspace{-3mm}
\paragraph{Noise injection.} We perform noise injection in two different ways, as described in the main paper. In \textit{GD (SGD) +noise} at each iteration we perturbe all the network weights with Gaussian noise with standard deviation $\sigma/\sqrt{M}$,where $M$ is the number of layers. Instead, in \textit{GD (SGD) +noise (layer)}, at each iteration, a specific layer\footnote{In the implementation we injected noise sequentially through the network, and we actually perturbe separately each parameter group~(i.e. half iterations noise is injected on specific layer biases) and pick $M$ to be the number of parameter groups.} is picked and all weights are perturbed with Gaussian noise with standard deviation $\sigma$. As explained in the main text~(see Section~\ref{sec:layer_wise_linear}\&\ref{sec:neural-net-deep}), we expect the layer-wise perturbation to provide a more stable approach for the minimization of the regularized loss, as width increases. In the results below, we show that this is indeed the case and provide detailed comparisons.

\subsection{Experiments on Fashion MNIST MLPs}

\begin{figure}
    \centering
    \includegraphics[height = 0.23\textwidth]{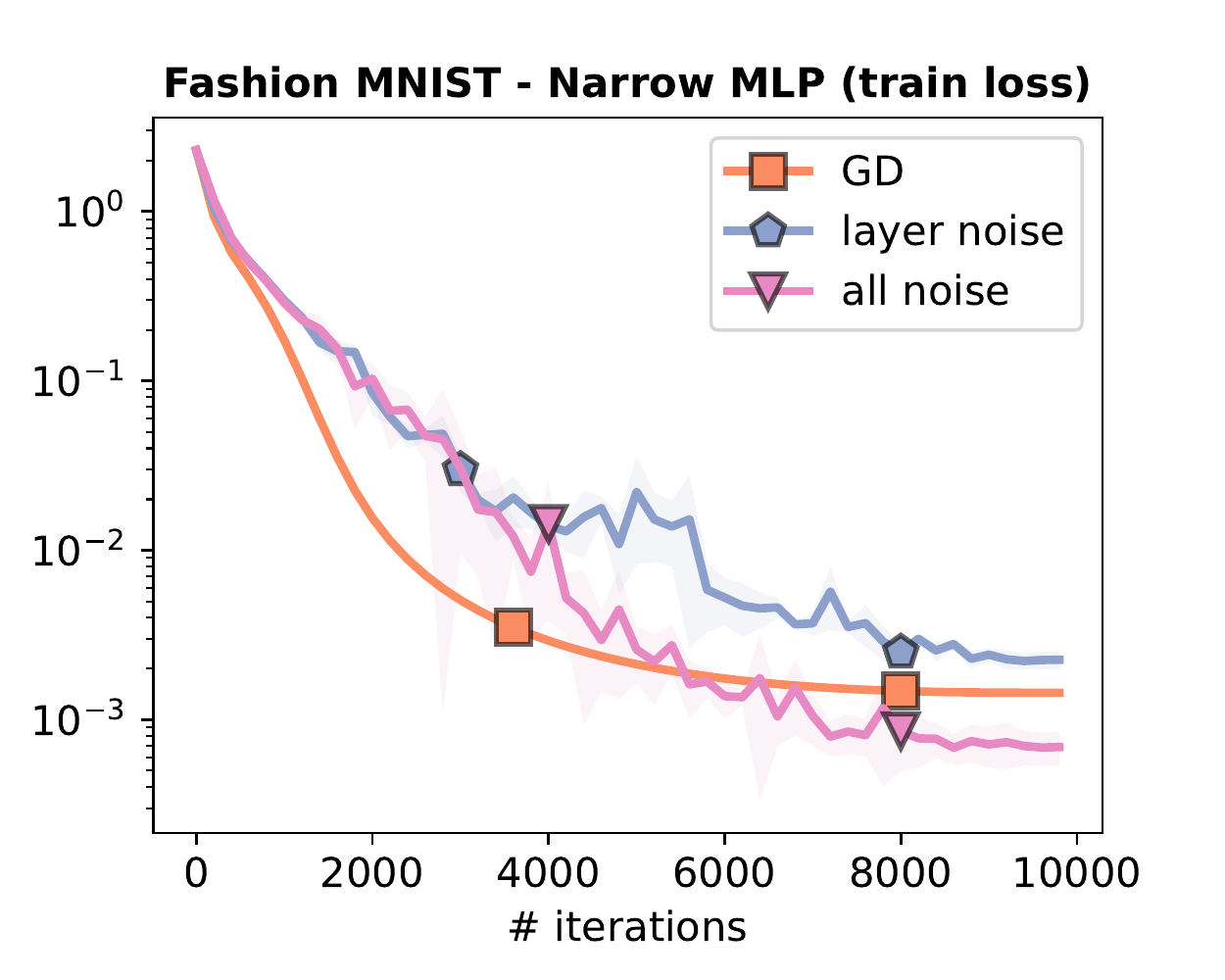}
    \includegraphics[height = 0.23\textwidth]{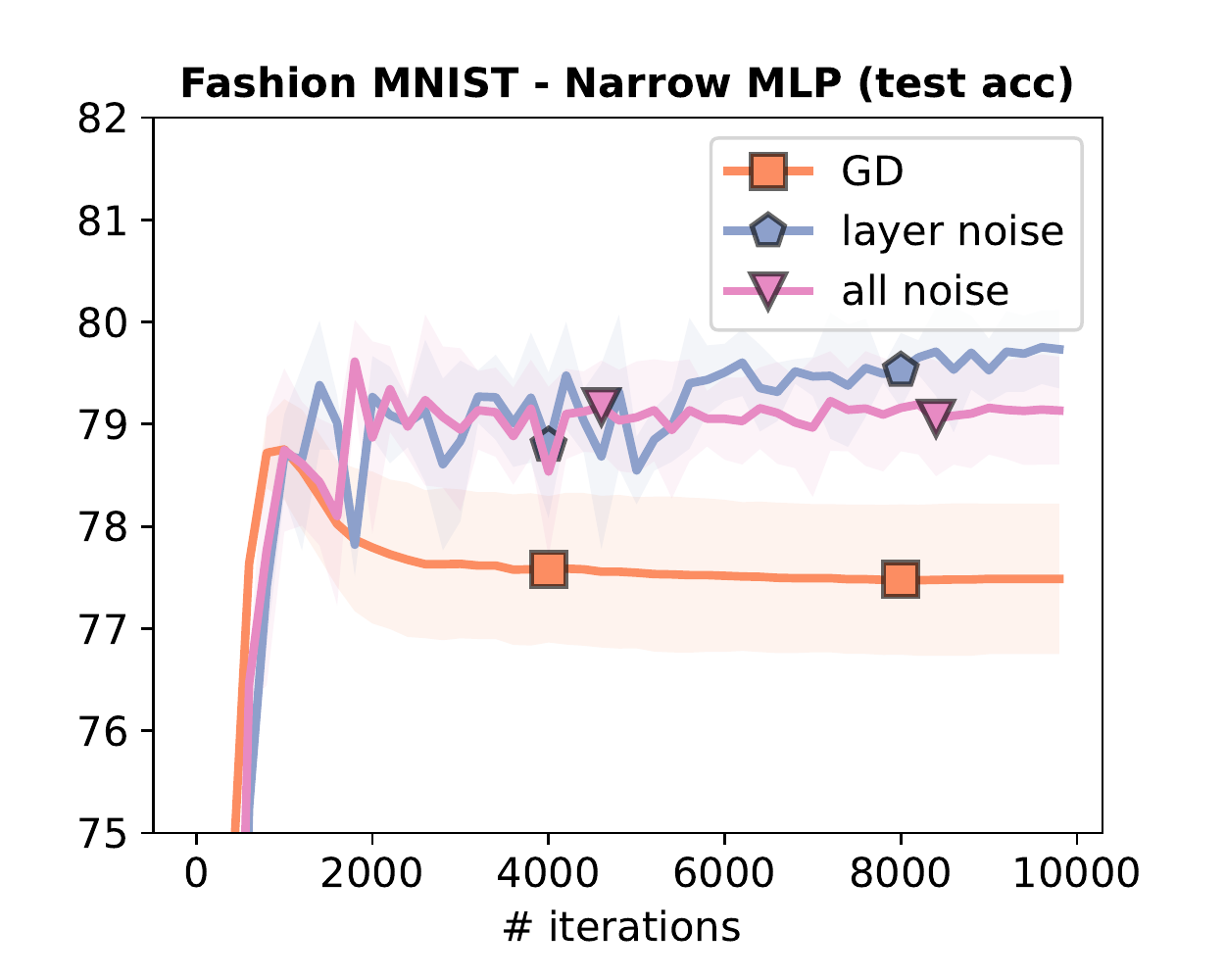}
    \includegraphics[height = 0.23\textwidth]{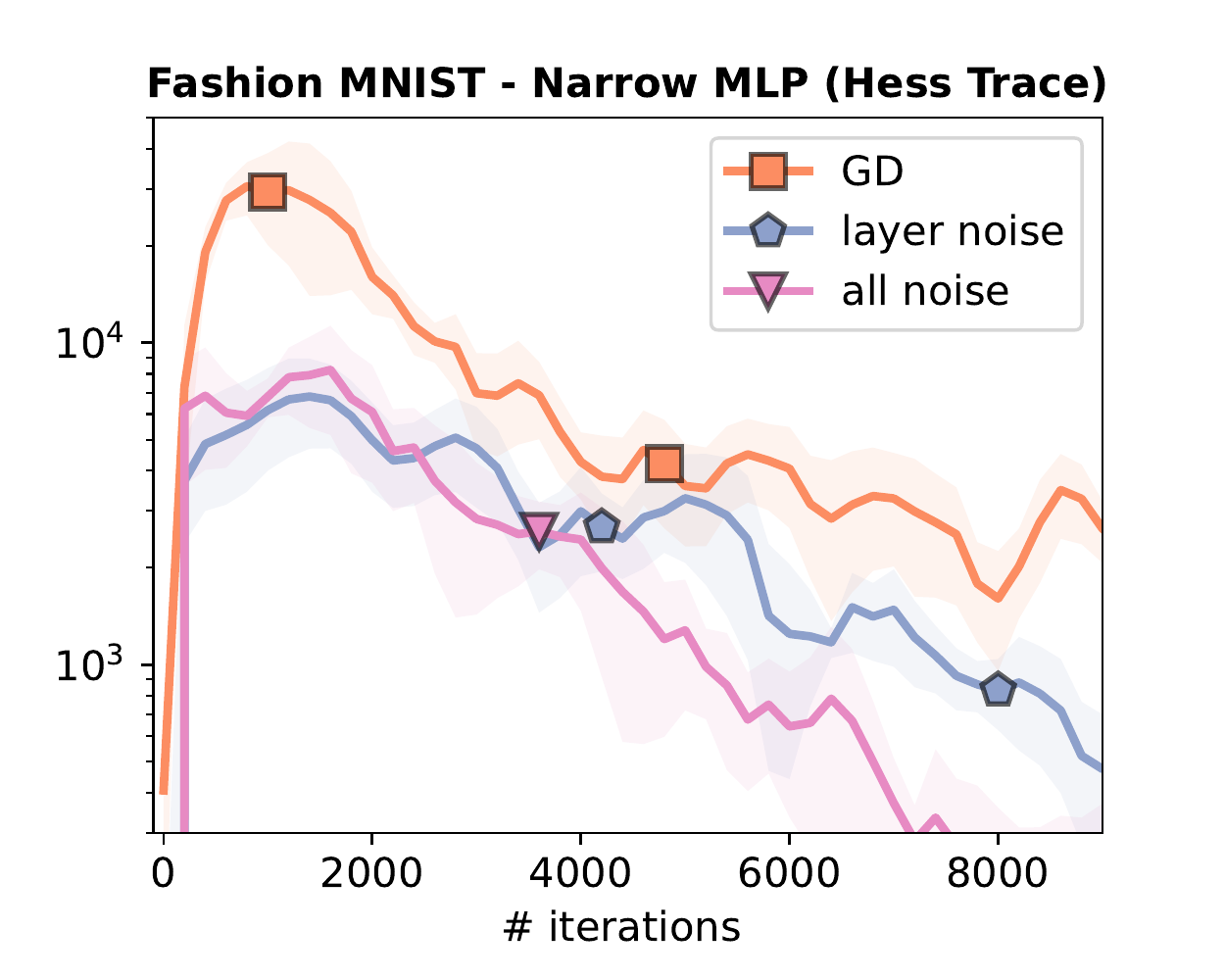}
    \vspace{-3mm}
    \caption{Fashion MNIST MLP 1: MLP with 4 Layers (3 hidden), width 500. Full Batch.}
    \vspace{-2mm}
    \label{fig:FMNIST1_app}
\end{figure}

\paragraph{Batch-size, learning rate and noise injection strength.} For the simulations on Fashion MNIST, we use a random subsample of the dataset ($1024$ samples), to introduce a sizable train-test gap. This also allows us to train with full-batch gradient descent on this subset. Performance of stochastic gradient descent~(i.e. mini-batch case) is evaluated for CNNs in the next subsection. We selected for all experiments a sizeable learning rate of $5e-3$, except for the Deep and Wide case~(MLP 4), where a smaller learning rate is necessary to avoid instabilities. Since we have no access to the problem Lipschitz constant, to allow methods to converge to potentially sharp regions, we use cosine annealing~\citep{loshchilov2016sgdr} for the learning rate. Regarding $\sigma$, we select the value that showcases best the implications of our theory: picking a value around $0.05$ works in all settings. A plot showing the effect of tuning $\sigma$ on both noise injection schemes can be found in Section~\ref{sec:exp} in the main paper. Test performance is evaluated on the full $10K$ testing dataset.
\vspace{-3mm}
\paragraph{Fashion MNIST MLP 1 (Shallow Narrow).} The input~($28\times 28 = 784$-dimensional) is processed with the following 4 narrow layers  with ReLU activations~(except last layer) to an output with dimension $10$:
\begin{equation}
    784 \rightarrow 500 \rightarrow 500 \rightarrow 500 \rightarrow 10.
\end{equation}
The final output is then processed using a log-softmax and signal is backpropagated using a cross-entropy loss. We run full-batch gradient descent on the 1024 datapoints~(subset of the data) and train with learning rate $5e-3$ and $\sigma = 5e-2$. Figure~\ref{fig:FMNIST1_app} shows that, since the network is not too wide, both the noise injection schemes are able to decrease the Hessian trace and improve test accuracy. Yet, interestingly, layer-wise noise injection outperforms standard noise injection in test accuracy.

\begin{figure}
    \centering
    \includegraphics[height = 0.23\textwidth]{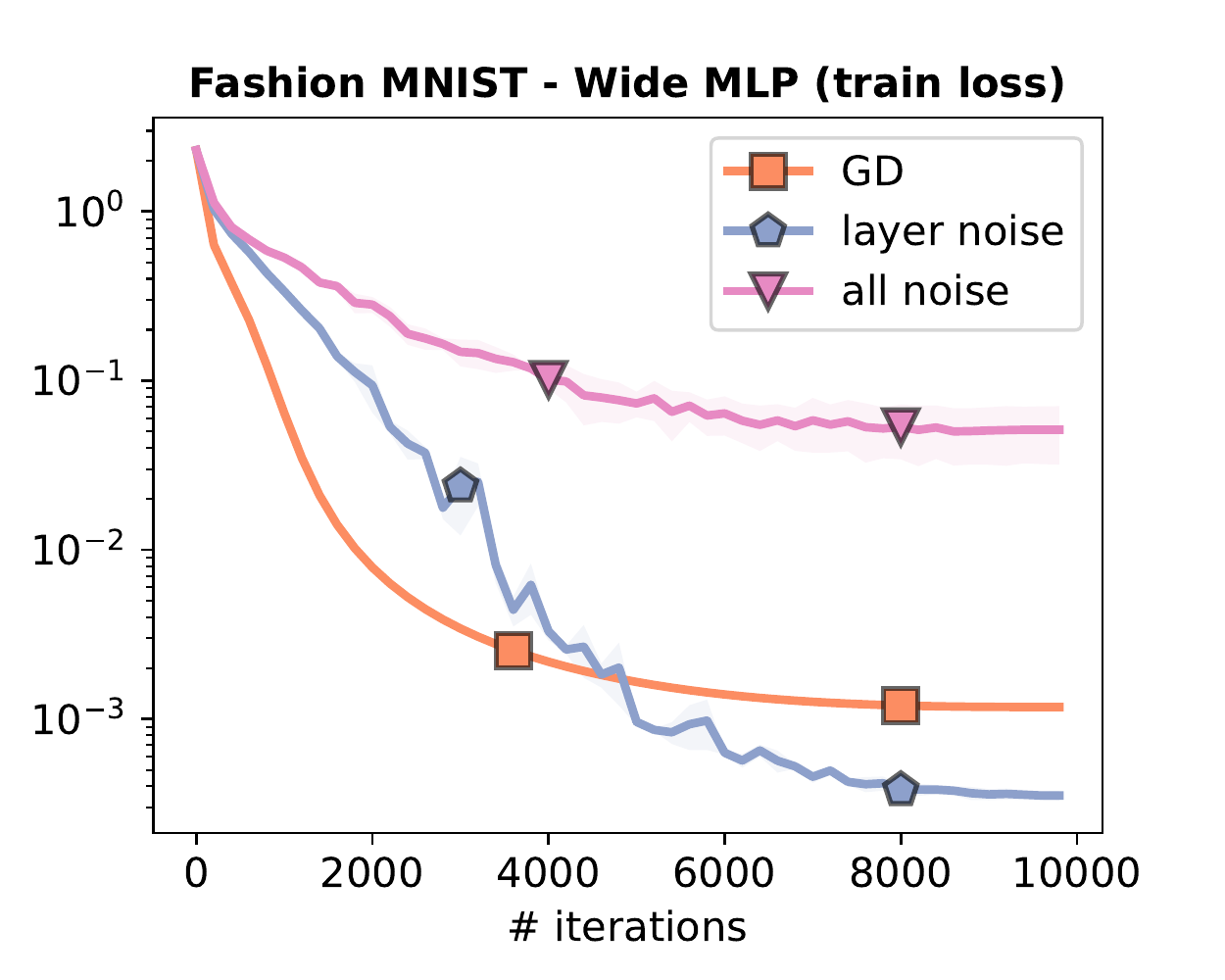}
    \includegraphics[height = 0.23\textwidth]{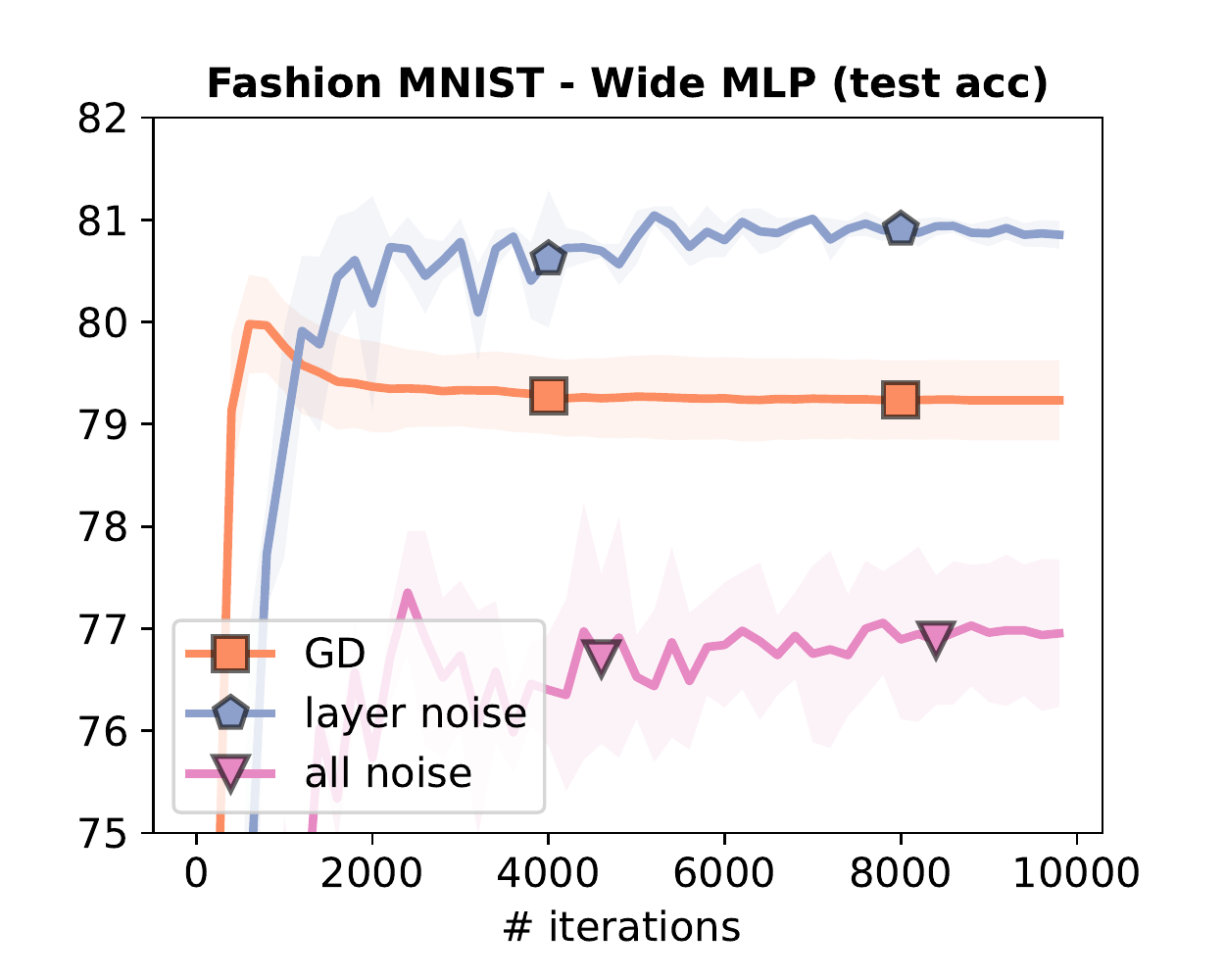}
    \includegraphics[height = 0.23\textwidth]{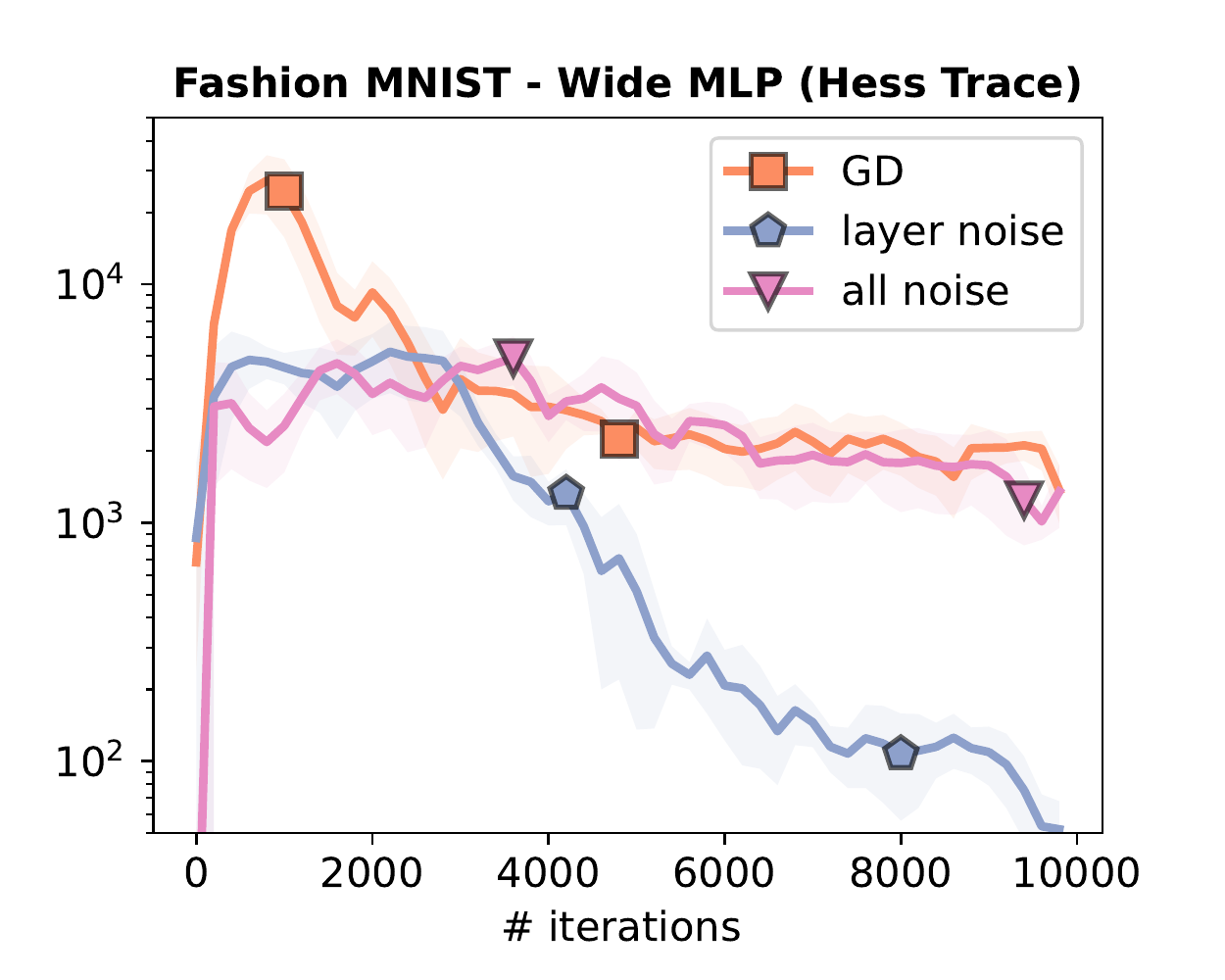}
    \vspace{-3mm}
    \caption{Fashion MNIST MLP 2: MLP with 4 Layers (3 hidden), width 5000. Full Batch.}
    \vspace{-2mm}
    \label{fig:FMNIST2_app}
\end{figure}

    \vspace{-3mm}
\paragraph{Fashion MNIST MLP 2 (Shallow Wide).} The $784$-dimensional input is processed with the following 4 wide layers with ReLU activations~(except last layer) to an output with dimension $10$:
\begin{equation}
    784 \rightarrow 5000 \rightarrow 5000 \rightarrow 5000 \rightarrow 10.
\end{equation}
The final output is then processed as before, and we select a stepsize $5e-3$ and $\sigma = 1e-1$. Figure~\ref{fig:FMNIST2_app} shows that, since the network is wide, only layer-wise injection provides a successful regularization. These findings are complemented by Figure~\ref{fig:tuning} in the main paper.

\begin{figure}
\vspace{-3mm}
    \centering
    \includegraphics[height = 0.23\textwidth]{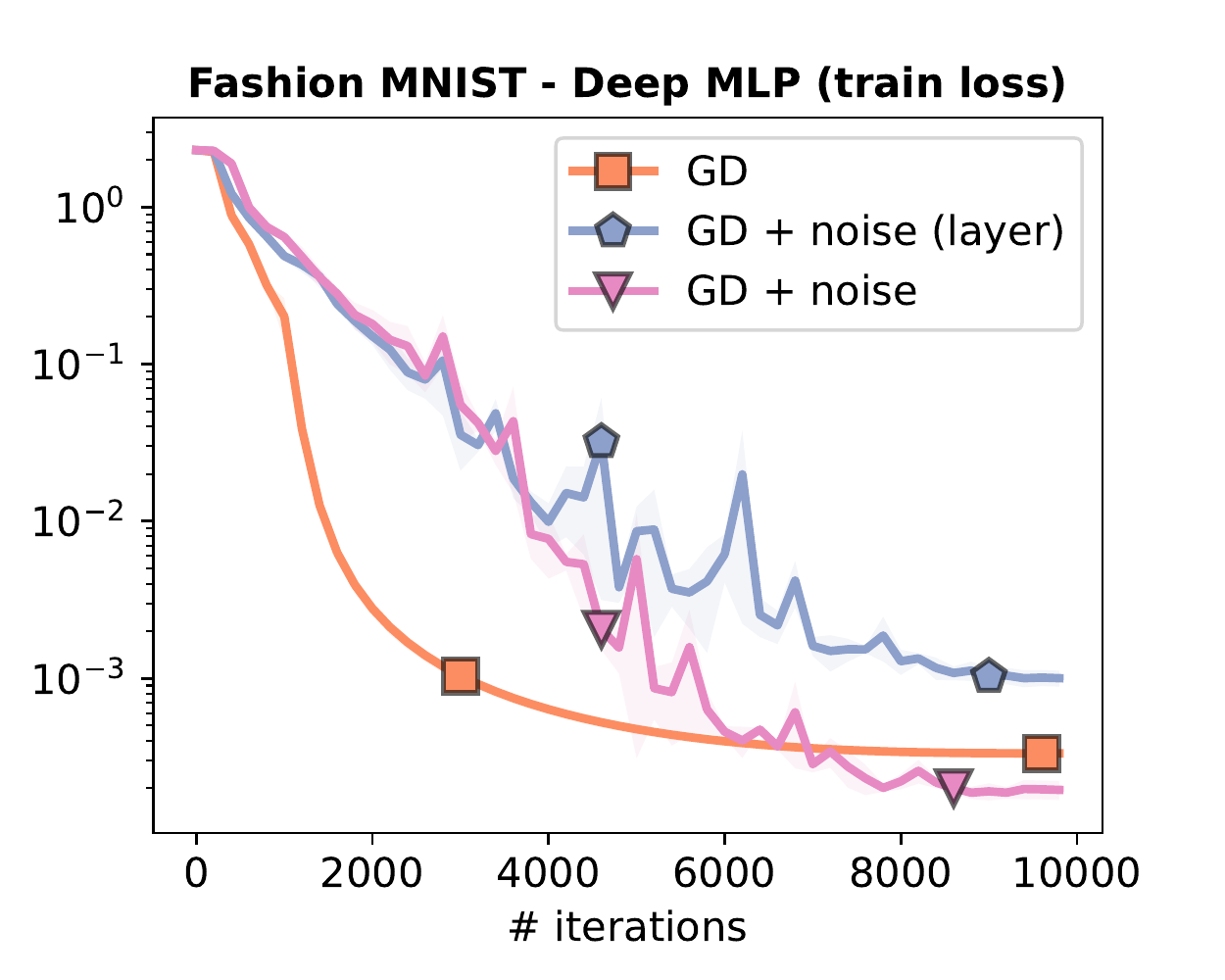}
    \includegraphics[height = 0.23\textwidth]{img/FMNIST_MLP3_test.pdf}
    \includegraphics[height = 0.23\textwidth]{img/FMNIST_MLP3_hess.pdf}
    \vspace{-3mm}
    \caption{Fashion MNIST MLP 3: MLP with 6 Layers (5 hidden), width 1000. Full Batch.}
    \vspace{-2mm}
    \label{fig:FMNIST3_app}
\end{figure}

\begin{figure}
    \centering
    \includegraphics[height = 0.23\textwidth]{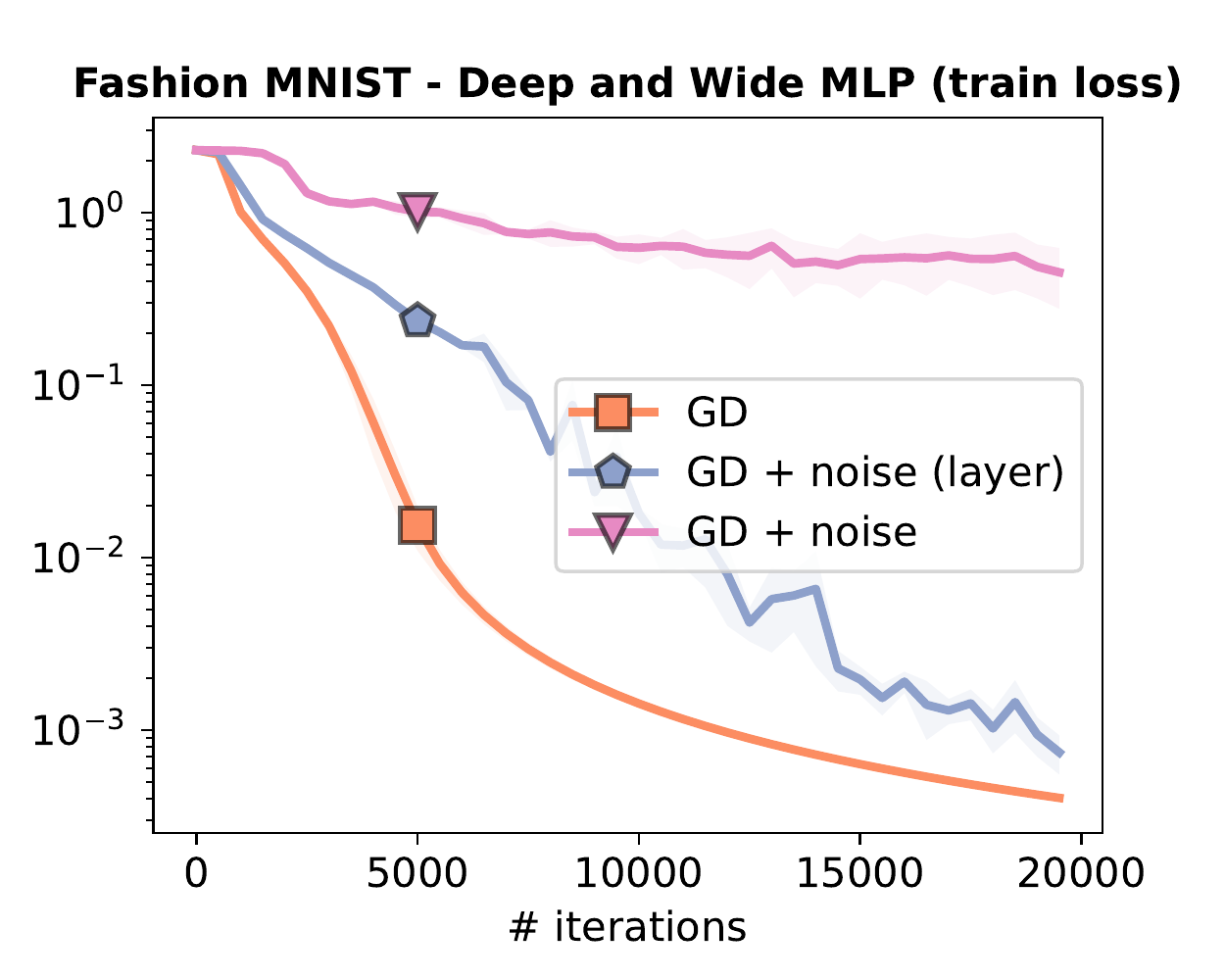}
    \includegraphics[height = 0.23\textwidth]{img/FMNIST_MLP4_test.pdf}
    \includegraphics[height = 0.23\textwidth]{img/FMNIST_MLP4_hess.pdf}
        \vspace{-3mm}
    \caption{Fashion MNIST MLP 4: MLP with 6 Layers (5 hidden), width 5000. Full Batch.}
    \vspace{-2mm}
    \label{fig:FMNIST4_app}
\end{figure}

    \vspace{-3mm}
\paragraph{Fashion MNIST MLP 3 (Deep Narrow).} $784$-dimensional input is processed with the following 6 not extremely wide layers with ReLU activations~(except last layer) to an output with dimension $10$:
\begin{equation}
    784 \rightarrow 1000 \rightarrow 1000 \rightarrow 1000 \rightarrow 1000 \rightarrow 1000 \rightarrow 10.
\end{equation}
The final output is then processed as before, and we select a stepsize $5e-3$ and $\sigma = 1e-1$. Figure~\ref{fig:FMNIST3_app} shows that the behavior is similar to the shallow case~(Figure~\ref{fig:FMNIST1_app}).
    \vspace{-3mm}
\paragraph{Fashion MNIST MLP 4 (Deep Wide).}
$784$-dimensional input is processed with the following 6 wide layers with ReLU activations~(except last layer) to an output with dimension $10$:
\begin{equation}
    784 \rightarrow 5000 \rightarrow 5000 \rightarrow 5000 \rightarrow 5000 \rightarrow 5000 \rightarrow 10.
\end{equation}
The final output is then processed as before, and we select a stepsize $1e-3$ and $\sigma = 5e-2$. Figure~\ref{fig:FMNIST4_app} shows that the behavior is similar to the narrow case~(Figure~\ref{fig:FMNIST2_app}): as the width increases, only layer-wise injection provides an improvement in test accuracy, while standard noise injection hurts convergence at the same value of $\sigma$~(scaled properly).

\subsection{Experiments on CIFAR10 CNNs}

\begin{figure}[ht]
    \centering
    \includegraphics[height = 0.23\textwidth]{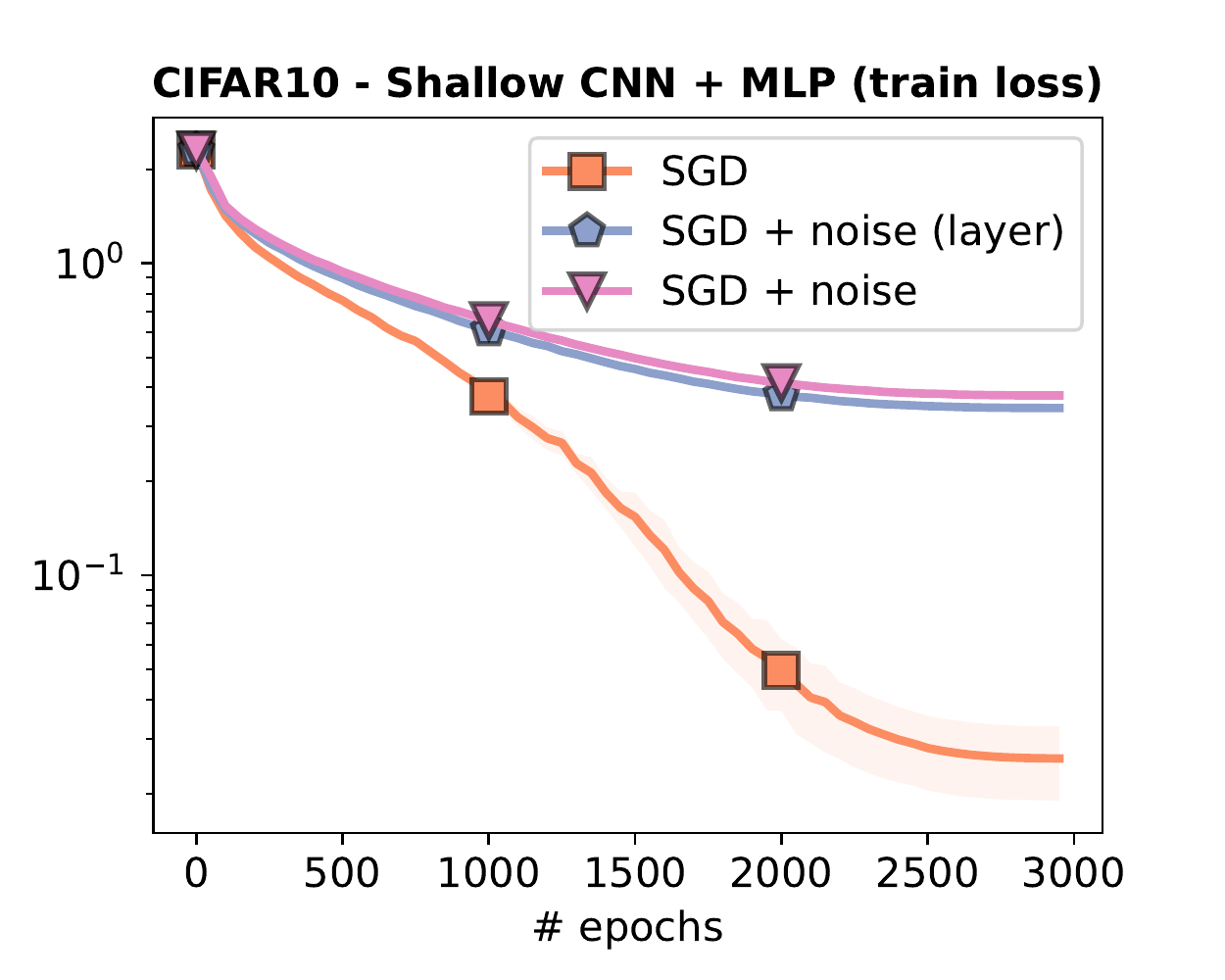}
    \includegraphics[height = 0.23\textwidth]{img/CIFAR10_CNN1_test.pdf}
    \includegraphics[height = 0.23\textwidth]{img/CIFAR10_CNN1_hess.pdf}
    \vspace{-3mm}
    \caption{CIFAR10, PyTorch Toy CNN+MLP. Batch size 1024.}
    \vspace{-2mm}
    \label{fig:CIFAR1}
\end{figure}

\paragraph{Batch-size and learning rate, noise injection strength.} For the simulations on CIFAR10, we use the full training dataset~($50K$ samples) and train with SGD with batch-size 1024 and cosine annealing on a tuned learning rate. $\sigma$ is picked to be $0.05$ in all settings and layer-wise noise injection is performed with $\sqrt{M}$ scaling compared to the standard case, where $M$ is the number of parameter groups in the network.  Test performance is evaluated on the full $10K$ testing dataset.
    \vspace{-3mm}
\paragraph{CIFAR10 Toy Narrow CNN+MLP.} The $28\times 28\times 3$ image~(RGB) is processed by 2 convolutions with 5 channels followed by an MLP. Both Conv and Lin layers have ReLU activations~(except last layer) and max pooling is applied after each convolution after the ReLU. Flattening of the tensor is performed before the MLP. The layers are reported\footnote{by Conv$(C_{in},C_{out}, k)$ we meen a convolutional layer with kernel size $k$, number of input channels $C_{in}$ and number output channels $C_{out}$} next:
\begin{equation}
         \text{Conv}(3, 6, 5), \text{Conv}(6, 16, 5), \text{Lin}(16 \times 5 \times 5, 120), \text{Lin}(120, 84), \text{Lin}(84, 10).
\end{equation}
Backpropagated gradients are computed using a cross-entropy loss. For this network, we train with SGD with batch-size 1024, using a stepsize of $1e-3$ with cosine annealing. We select here $\sigma = 5e-2$. In Figure~\ref{fig:CIFAR1} we see that, while the test performance of SGD with batch size 1024 degrades overtime~(known effect of linear layers), the performance of all noise injection schemes leads to better and stable generalization. This is reflected in the regularized Hessian trace. We note that here the performances of the noise injection methods are comparable since the network is not very wide.
    \vspace{-3mm}
\paragraph{CIFAR10 Wider fully convolutional.}
 The $28\times 28\times 3$ image~(RGB) is processed by a fully convolutional network with up to 128 channels~(i.e. wide) with max pooling and ReLU activations~(except last layer). The last layer acts on the average output of each channel. The network operations are as follows:
\begin{equation}
\text{Conv}(3, 32, 3), \text{Conv}(32, 64, 3), \text{Conv}(64, 128, 3), \text{Conv}(128, 128, 3), \text{Lin}(128, 10)
\end{equation}

\begin{figure}[ht]
    \centering
    \includegraphics[height = 0.23\textwidth]{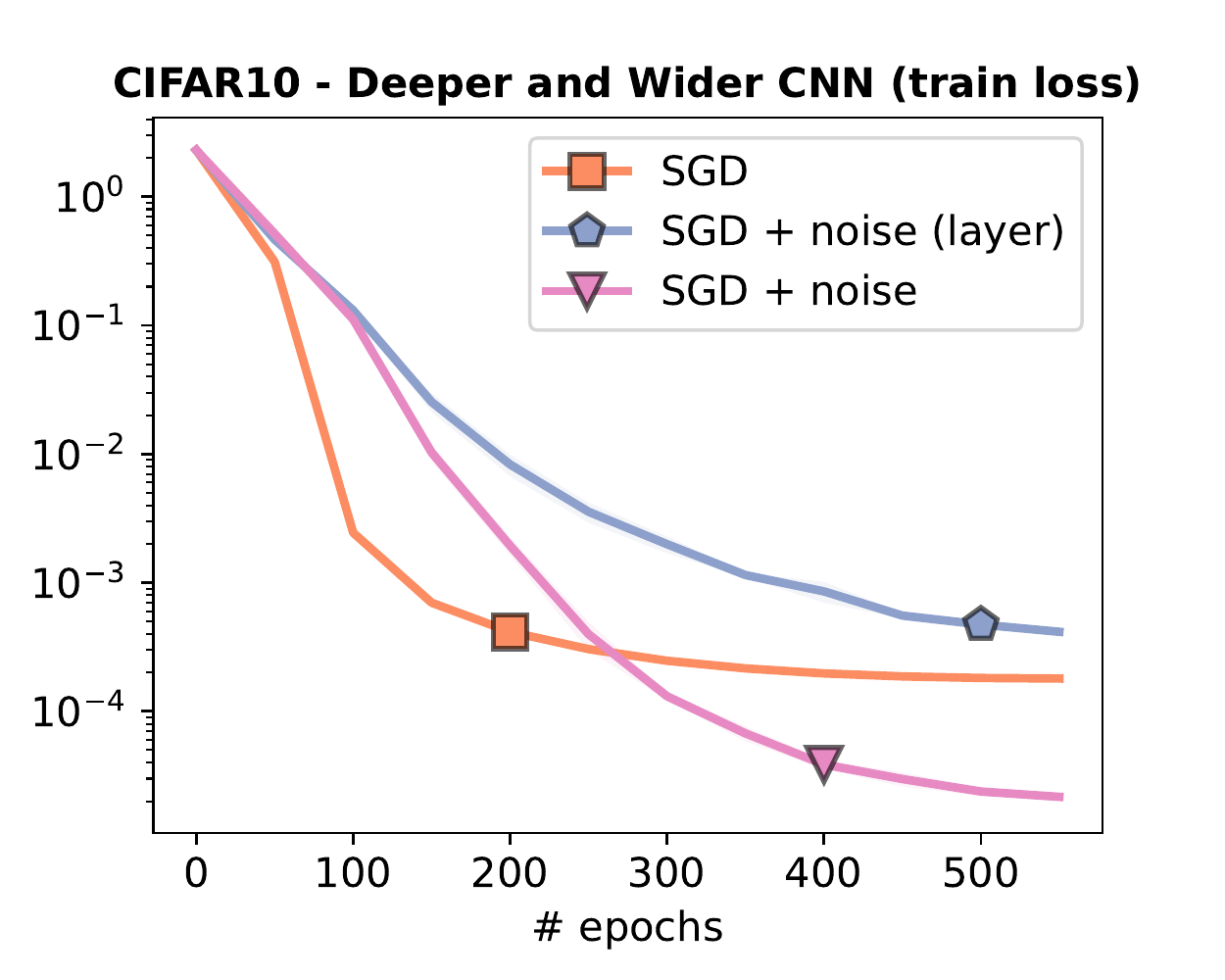}
    \includegraphics[height = 0.23\textwidth]{img/CIFAR10_CNN2_test.pdf}
    \includegraphics[height = 0.23\textwidth]{img/CIFAR10_CNN2_hess.pdf}
    \vspace{-3mm}
    \caption{CIFAR10, Wide CNN (128 channels). Batch size 1024.}
    \vspace{-2mm}
    \label{fig:CIFAR2}
\end{figure}

For this network, we train with SGD with batch-size 1024, using a stepsize of $5e-3$ with cosine annealing. We select here $\sigma = 5e-2$. Figure~\ref{fig:CIFAR2} shows that, in contrast to the toy narrow CNN we inspected above, here layer-wise noise injection provides an improved performance in terms of test loss over standard noise injection. Yet, compared to the MLP case, here injecting noise to all weights does not lead to a quick performance degradation --- to observe this, one has to design a network with unrealistically big number of channels. Regarding the Hessian trace, we observe that injecting noise dampens the oscillations in the dynamics of this quantity, which is in line with our result on minimization of a regularized loss~(see Section~\ref{sec:layer_wise_linear}), even though this was derived in the MLP case.

\end{document}